\documentclass[sigconf, nonacm]{acmart}
\setcopyright{none}           
\renewcommand\acmConference[4][]{} 
\renewcommand{\acmDOI}{}      
\renewcommand{\acmISBN}{}     
\acmSubmissionID{XXXX}        
\usepackage{enumitem}
\usepackage{amsmath}
\usepackage{graphicx}

\usepackage{booktabs} 
\usepackage{multirow} 
\usepackage{graphicx} 
\usepackage{xcolor}   
\usepackage[table]{xcolor}

\usepackage{geometry}
\usepackage{booktabs}
\usepackage{tabularx}

\definecolor{correctgreen}{RGB}{46, 139, 87}   
\definecolor{wrongred}{RGB}{178, 34, 34}       
\definecolor{evidenceblue}{RGB}{30, 144, 255}  
\definecolor{hintorange}{RGB}{255, 140, 0}     
\definecolor{taggray}{RGB}{105, 105, 105}      

\begin{document}

\title{SIGHT: Reinforcement Learning with Self-Evidence and Information-Gain Diverse Branching for Search Agent}



\author{Wenlin Zhong}
\affiliation{%
  \institution{Zhejiang University}
  \city{Hangzhou}
  \country{China}}
\email{zhongwenlin@zju.edu.cn}

\author{Jinluan Yang}
\affiliation{%
  \institution{Zhejiang University}
  \city{Hangzhou}
  \country{China}}
\email{yangjinluan@zju.edu.cn}

\author{Yiquan Wu}
\affiliation{%
  \institution{Zhejiang University}
  \city{Hangzhou}
  \country{China}}
\email{wuyiquan@zju.edu.cn}

\author{Yi Liu}
\affiliation{%
  \institution{Chongqing Ant Consumer Finance Co., Ltd.}
  \city{Chongqing}
  \country{China}}
\email{larry.liuy@myxiaojin.cn}

\author{Jianhang Yao}
\affiliation{%
  \institution{Alibaba Group}
  \city{Hangzhou}
  \country{China}}
\email{jianhang.yjh@alibaba-inc.com}

\author{Kun Kuang}
\affiliation{%
  \institution{Zhejiang University}
  \city{Hangzhou}
  \country{China}}
\email{kunkuang@cs.zju.edu.cn}

\renewcommand{\shortauthors}{Zhong et al.}

\begin{abstract}

Reinforcement Learning (RL) has empowered Large Language Models (LLMs) to master autonomous search for complex question answering. However, particularly within multi-turn search scenarios, this interaction introduces a critical challenge: search results often suffer from high redundancy and low signal-to-noise ratios. Consequently, agents easily fall into "Tunnel Vision," where the forced interpretation of early noisy retrievals leads to irreversible error accumulation. To address these challenges, we propose SIGHT, a framework that enhances search-based reasoning through Self-Evidence Support (SES) and Information-Gain Driven Diverse Branching. SIGHT distills search results into high-fidelity evidence via SES and calculates an Information Gain score to pinpoint pivotal states where observations maximally reduce uncertainty. This score guides \textit{Dynamic Prompting Interventions}—including \textit{de-duplication}, \textit{reflection}, or \textit{adaptive branching}—to spawn new branches with SES. Finally, by integrating SES and correctness rewards via Group Relative Policy Optimization, SIGHT internalizes robust exploration strategies without external verifiers.
Experiments on single-hop and multi-hop QA benchmarks demonstrate that SIGHT significantly outperforms existing approaches, particularly in complex reasoning scenarios, using fewer search steps.


\end{abstract}




\maketitle

\section{Introduction}
Large Language Models (LLMs) have evolved from static Retrieval-Augmented Generation (RAG) to dynamic Agentic Workflows like ReAct \cite{yao2022react} to address knowledge-intensive tasks \cite{gao2023retrieval,li2025search}. However, constrained by the fragility of prompt-driven in-context learning, the field has pivoted towards Agentic Reinforcement Learning (Agentic RL) \cite{jin2025search}. By shifting training to dynamic agent-environment reasoning via outcome-based rewards (e.g., GRPO \cite{shao2024deepseekmath}), Agentic RL enables models to master autonomous search. Crucially, solving complex queries in this paradigm entails multi-step reasoning interleaved with real-time feedback \cite{zhang2025deep}. This sequential nature necessitates exploring effective \textit{step-level} search behaviors, rather than merely optimizing the trajectory based on sparse outcomes.

\begin{figure}[t!]
    \centering
    \includegraphics[width=1\linewidth]{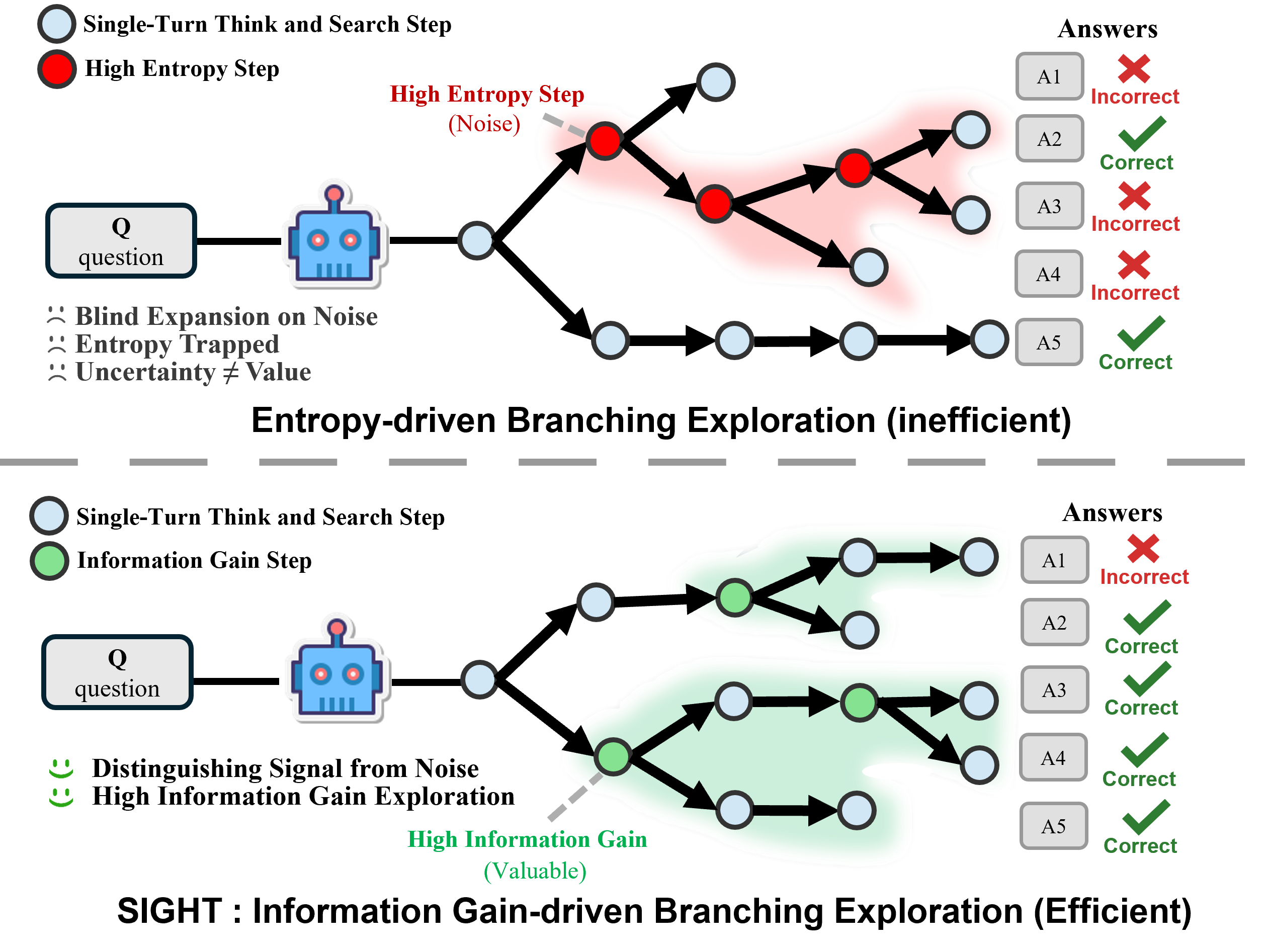}
    \caption{Contrast between (top) noise-sensitive entropy exploration and (bottom) SIGHT's noise-resilient, Information Gain-driven framework.}
    \label{fig:intro}
\end{figure}

To encourage such step-level exploration, recent studies have utilized heuristic branching based on randomness or entropy \cite{ji2025tree, dong2025agentic, shen2026carlfocusingagenticreinforcement, li2025parallelmuseagenticparallelthinking}. While promoting diversity, these methods implicitly rely on a fragile assumption: that environmental feedback is intrinsically clean. They overlook the volatility of real-world environments, where raw search results suffer from severe redundancy and low signal-to-noise ratio (SNR). Consequently, indiscriminately expanding search branches based on uncertainty metrics triggers \textbf{Tunnel Vision}, where imperfect initial steps lock the model into a suboptimal path \cite{wen2025parathinker}. This challenge is acutely amplified in search agents \cite{xinmiao2026webanchor}: a misguided query yields noisy observations, forcing the agent to hallucinate connections in flawed contexts rather than backtracking, leading to irreversible error accumulation.

As illustrated in Figure~\ref{fig:intro}, we contrast heuristic methods (e.g., entropy-based) with our proposed approach. In the top panel, heuristic agents mistake noisy steps for valuable uncertainty, triggering indiscriminate branching on noise-ridden contexts (red flow) that leads to \textit{Tunnel Vision}. \textbf{This necessitates a noise-aware framework capable of discriminating valid feedback from stochastic interference.} To this end, we propose \textbf{SIGHT}, a novel Agentic RL framework that enhances search-based reasoning through \textbf{Self-Evidence Support (SES)} and \textbf{Information-Gain Driven Diverse Branching}. Conversely (bottom panel), SIGHT leverages Information Gain to filter distractions, restricting exploration to paths with genuine informational value (green flow) to efficiently yield higher-quality rollouts.

To implement this, SIGHT first introduces SES steps to distill raw search results into noise-free evidence, acting as a preliminary filter. To optimize \textit{when} and \textit{how} to explore, we calculate an Information-Gain (IG) score—defined as the posterior-prior likelihood difference—to quantify the contribution of tool results toward answer certainty. A high IG score identifies pivotal states containing critical evidence, justifying adaptive branching, while a low score signals potential "noise traps," triggering corrective exploration. Based on this score, we introduce Dynamic Prompting Interventions. Instead of branching via naive state duplication, SIGHT injects targeted interventions to enforce semantic diversity: specifically, it injects De-duplication prompts to avoid search query redundancy, employs Reflective prompts to rectify linear reasoning errors, or triggers Adaptive Branching at pivotal states, spawning new paths rooted in SES-distilled evidence.
Finally, we incorporate these mechanisms and SES rewards alongside answer correctness rewards using Group Relative Policy Optimization (GRPO). This enables the model to internalize robust exploration strategies without relying on an external verifier. Experiments on single-hop and multi-hop QA benchmarks demonstrate that SIGHT significantly outperforms existing approaches, particularly in complex reasoning scenarios, while requiring fewer search steps.

Our main contributions are summarized as follows:
\begin{itemize}
\item We explore the Tunnel Vision problem in multi-turn search agents, revealing how noisy retrievals lead to irreversible reasoning collapses in trajectory-level RL.
\item We propose SIGHT, a novel framework that integrates Self-Evidence Support (SES) and Information-Gain (IG) Driven Diverse Branching. By leveraging IG scores to optimize \textit{when} to branch (distinguishing pivotal states) and deploying Dynamic Prompting Interventions (e.g., De-duplication, Reflection) to guide \textit{how} to branch, ultimately achieves robust, information-gain driven branching exploration.
\item We demonstrate through extensive experiments on single-hop and multi-hop QA benchmarks that SIGHT significantly outperforms existing approaches. Notably, it achieves superior accuracy with fewer search steps, proving high computational efficiency and sample robustness.
\end{itemize}

\section{Related Work}

\noindent \textbf{Reinforcement Learning for LLM Reasoning.}
Reinforcement Learning has emerged as a pivotal paradigm for enhancing the reasoning capabilities of large language models, addressing the drawbacks of static imitation from previous supervised fine-tuning \cite{guo2025deepseek,cao2026pushing}. While initial efforts like RLHF with PPO~\cite{schulman2017proximal} aligned models with human preferences, the necessity of maintaining a concurrent value model (critic) introduces substantial memory overhead and training instability. Consequently, \textit{Outcome-based RL} paradigms have emerged, from Group Relative Policy Optimization (GRPO) \cite{shao2024deepseekmath} to diverse policy variants such as DAPO \cite{yu2025dapo}, GSPO \cite{zheng2025group}, SAPO \cite{gao2025softadaptivepolicyoptimization} and Tree-GRPO \cite{ji2025tree}. 
Different from previous works that focus on utilizing the entropy signal across the whole training stage to optimize the rollout strategies for better exploration, our work especially mitigates the Tunnel Vision \cite{wen2025parathinker,xinmiao2026webanchor} issue from early noisy exploration and proposes the information gain-driven exploration strategy to further achieve efficient and effective exploration.    \\

\noindent \textbf{Reinforcement Learning for Search Agent.} To equip large language models with knowledge beyond their internal parameters, a significant line of research focuses on augmenting them with the ability to invoke external tools (e.g., search engines) during reasoning \cite{zhang2025deep,team2025tongyi}. Representative work, such as Search-R1 \cite{jin2025search, he2026searchr2enhancingsearchintegratedreasoning, li2025websailornavigatingsuperhumanreasoning, shi2025searchrefinethinkfacilitating, qian2025toolrlrewardtoollearning,wang2025actingreasoningmoreteaching,chen2025reinforcement,zhou2026lras}, frames the LLM as a search agent that interleaves reasoning with search in a multi-turn interactive environment. Building upon this "search-during-think" paradigm, more challenging scenarios involving complex queries have garnered increasing attention. This is because such scenarios necessitate effective step-level exploration, going beyond reliance on sparse terminal outcomes \cite{zheng2025deepresearcher}. Recent studies, such as ARPO \cite{dong2025agentic}, employ heuristic branching based on entropy, perplexity, or randomness to advance the step-level exploration strategies \cite{li2025parallelmuseagenticparallelthinking,ji2025tree}. While these strategies promote diversity, they typically neglect the adverse effects of real-world retrieval noise. Indiscriminately expanding branches on such feedback triggers Tunnel Vision, locking agents into suboptimal paths \cite{xinmiao2026webanchor}. This motivates SIGHT, which actively discriminates valid feedback from noise to achieve robust, information-gain driven exploration.


\section{Preliminaries}


\subsection{Multi-Turn Agent Reasoning}
We follow the standard ReAct-style agentic workflow~\cite{yao2022react}. Unlike static single-turn interaction, the agent engages in multi-turn Thought-Action-Observation cycles with the environment to solve a given task. Specifically, at each step $t = 0, 1, ..., T - 1$, the LLM generates a thought $\tau_t$ and a parsable textual action $\alpha_t$ based on the existing context $s_t$. The action typically corresponds to tool use, through which the agent dynamically interacts with the environment to obtain new observations $o_t$. A complete $T$-step agent rollout consists of three interleaved trajectories:
\begin{equation}
    \mathcal{H}_{T} = \{(\tau_0, \alpha_0, o_0), (\tau_1, \alpha_1, o_1), ..., (\tau_{T-1}, \alpha_{T-1}, o_{T-1}),   a_{T+1}\}.
\end{equation}

where $\tau_t$ represents the reasoning thought, $\alpha_t$ denotes the action taken by the agent (e.g., a search query), $o_t$ is the observation from the environment (or search engine), and $a_{T+1}$ denotes the final answer. At each step $t$, the policy model $\pi_\theta$ samples a thought-action pair based on the preceding history $\mathcal{H}_{t-1}$:
\begin{equation}
    (\tau_t, a_t) \sim \pi_\theta(\cdot \mid \mathcal{H}_{t-1}).
\end{equation}

Consistent with the formulation in~\cite{wang2025ragen}, we model this dynamic interaction as a Markov Decision Process (MDP). Process $\mathcal{M} = \{S, A, P\}$, where $S$ denotes the state space, corresponding to the accumulated interaction history $\mathcal{H}_{<t}$ prior to step $t$. The action space $A$ is composite, where each action consists of a reasoning thought paired with an environmental operation $(\tau_t, \alpha_t)$. $P$ represents the transition dynamics, governing both the stochastic environmental feedback $P_{\text{env}}$ and the deterministic context updates. The generative process is driven by the LLM policy $\pi_\theta$ as follows:

\begin{equation}
\begin{split}
    p_\theta&(s_{0:T}, \tau_{0:T}, \alpha_{0:T}, o_{0:T}) = \\
    & p(s_0) \prod_{t=0}^{T-1} \left[ \pi_\theta(\tau_t | s_t) \pi_\theta(\alpha_t | s_t, \tau_t) P_{\text{env}}(o_{t+1} | \alpha_t) \right].
\end{split}
\end{equation}

\subsection{Agentic Reinforcement Learning}

With the multi-turn interaction formalized as a Markov Decision Process (MDP), Reinforcement Learning (RL) can be directly applied to optimize the policy space by maximizing the expected return of the full state-action trajectory~\cite{wang2025ragen}.

To efficiently optimize this objective and enhance the reasoning capabilities of the agent within a dynamic environment, we adopt Group Relative Policy Optimization (GRPO)~\cite{shao2024deepseekmath}, extending it to support interleaved retrieval and reasoning. Consequently, the generation process is treated not as a static sequence but as a trajectory $\mathcal{H}$ resulting from the interaction between the policy and a search engine $\mathcal{R}$.

Given a question-answer pair $(q, a)$, the behavior policy $\pi_{\theta_{\text{old}}}$ interacts with $\mathcal{R}$ to sample a group of $G$ trajectories $\{\mathcal{H}_i\}_{i=1}^G$, where each trajectory $\mathcal{H}_i$ consists of a sequence of thoughts, actions (queries), observations, and the final answer. The objective function, incorporating the search engine $\mathcal{R}$ as part of the transition dynamics, updates model parameters $\theta$ as follows:

\begin{equation} \label{eq:agentic_grpo}
\begin{split}
    \mathcal{J}_{\text{GRPO}}(\theta) =& \mathbb{E}_{\substack{(q,a)\sim D, \{\mathcal{H}_i\}_{i=1}^G \sim \pi_{\theta_{\text{old}}}(\cdot|q; \mathcal{R})}} 
    \\ 
    & \Bigg[ \frac{1}{G} \sum_{i=1}^G \frac{1}{|\mathcal{H}_i|} \sum_{t=1}^{|\mathcal{H}_i|} \bigg\{ \min \bigg[ \frac{\pi_\theta(h_{i,t}|q, \mathcal{H}_{i,<t}; \mathcal{R})}{\pi_{\theta_{\text{old}}}(h_{i,t}|q, \mathcal{H}_{i,<t}; \mathcal{R})} \hat{A}_{i,t}, \\
    & \text{clip}\left(\frac{\pi_\theta(h_{i,t}|q, \mathcal{H}_{i,<t}; \mathcal{R})}{\pi_{\theta_{\text{old}}}(h_{i,t}|q, \mathcal{H}_{i,<t}; \mathcal{R})}, 1-\epsilon, 1+\epsilon\right) \hat{A}_{i,t} \bigg] \\
    & - \beta \mathbb{D}_{\text{KL}}\left(\pi_\theta \| \pi_{\text{ref}}\right) \bigg\} \Bigg]
\end{split}
\end{equation}

where $h_{i,t}$ represents the $t$-th token (or step) in the $i$-th trajectory. The advantage estimation $\hat{A}_{i,t}$ is computed by comparing the trajectory return against the group baseline derived from a set of $G$ trajectories $\{\mathcal{H}_i\}_{i=1}^G$ sampled for the same prompt. Specifically, given the outcome-based rewards $\{R_i\}_{i=1}^G$ corresponding to each trajectory $\mathcal{H}_i$ (e.g., correctness of the final answer after retrieval), the advantage is standardized as:

\begin{equation}
\label{eq:advantage_r}
\hat{A}_{i,t} = \frac{R_i - \text{mean}(\{R_i\}_{i=1}^G)}{\text{std}(\{R_i\}_{i=1}^G)}
\end{equation}

Here, the reward $R_i$ serves as a verifiable signal (e.g., $1$ for a correct answer, $0$ otherwise). This group-normalized objective steers the policy to assign higher probabilities to trajectories that outperform their peers within the same sampled batch. Crucially, by explicitly conditioning the update on the search engine $\mathcal{R}$, this formulation ensures that the model is optimized to jointly refine its internal reasoning and its effective utilization of external tools to maximize the final trajectory reward.

\section{Methodology}
\label{sec:methodology}

\begin{figure*}[t!]
    \centering
    \includegraphics[width=1\linewidth]{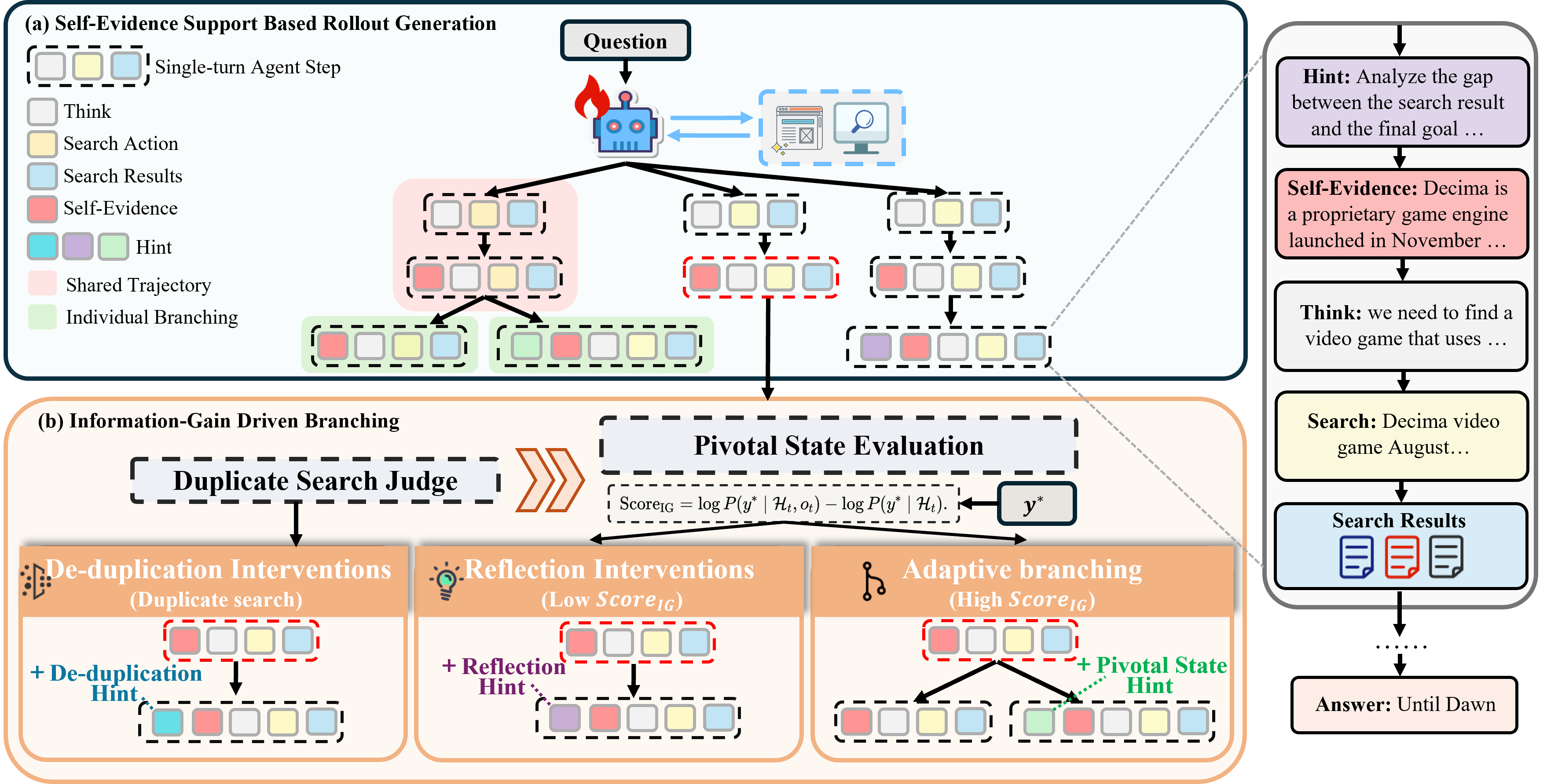}
    \caption{The Overall framework of SIGTHT. (a) Self-Evidence Support (SES) based rollout generation for active noise filtration: After every search action, the agent autonomously distills the raw observation $o_t$ (e.g., single-turn search result), into a noise-free evidence snippet $e_t$ within a \texttt{<self-evidence>} tag; (b) Information-Gain Driven Branching for Exploration: The system first performs pivotal state evalution to calculate an Information Gain (IG) score to quantify the value of the current state. Based on the IG score and interaction history, specific \textit{Dynamic Prompting Interventions} are injected to guide the next reasoning step—triggering de-duplication, reflection, or adaptive branching. SIGHT employs a continuous filtering and monitoring mechanism at each interaction step $t$. }
    \label{fig:main}
\end{figure*}

In this section, we present \textbf{SIGHT}, a noise-aware Agentic RL framework for multi-turn search tasks, designed to guide LLMs in exploring step-wise search behaviors under Information-Gain guidance.
As illustrated in Figure \ref{fig:main}, SIGHT enhances the standard agentic RL ``search-during-think'' paradigm by integrating two novel mechanisms: Self-Evidence Support (SES) based rollout generation for active noise filtration and Information-Gain Driven Diverse Branching for efficient exploration.



\subsection{SES-Based Rollout Generation}
\label{subsec:ses}

Raw search results often contain severe redundancy and irrelevant context (low Signal-to-Noise Ratio), which can distract the policy model and lead to hallucinations. To address this, we introduce \textbf{Self-Evidence Support (SES)}, a mechanism that acts as a preliminary filter applied in every interaction turn.\\

\noindent \textit{\textbf{SES Mechanism}}
Specifically, immediately after receiving a raw search result $o_t$ within a \texttt{<result>} block, the LLM is prompted to generate a \texttt{<self-evidence>} tag. Inside this tag, the model distills $o_t$ into concise, query-relevant evidence $e_t$, discarding noise:
\begin{equation}
    o_t \xrightarrow{\text{LLM}} \texttt{<self-evidence>} e_t \texttt{</self-evidence>}.
\end{equation}
The agent's subsequent "Think" step is then conditioned on this filtered evidence $e_t$ rather than the noisy raw observation. This ensures that the reasoning chain remains grounded in high-fidelity information.\\

\noindent \textit{\textbf{Rollout Generation}}
The actor LLM $\pi_\theta$ orchestrates the trajectory generation through iterative interactions with the search engine $\mathcal{R}$. As depicted in Figure~\ref{fig:main}, each interaction cycle encompasses a structured sequence of operations: reasoning via \texttt{<think>}, querying the search engine with \texttt{<search>}, ingesting retrieved documents through \texttt{<result>}, and crucially, purifying raw data into noise-free insights using \texttt{<self-evidence>}. Following multi-turn search interactions, once the model deems the accumulated context sufficient to answer the question, it synthesizes the final response within an \texttt{<answer>} block. Crucially, the model autonomously determines the number of search rounds, dynamically adapting the interaction depth based on both the problem complexity and the quality of the search results. Refer to Appendix~\ref{app:prompt_hint} for the system prompt.

 \subsection{Information-Gain Driven Diverse Branching}
\label{subsec:igdb}
While SES filters noise as stated above, it does not guarantee efficient exploration. To optimize \textbf{\textit{when}} and \textbf{\textit{how}} to explore, we propose the Information-Gain Driven Diverse Branching mechanism. This module evaluates the current state and applies Dynamic Prompting Interventions via hint tokens to steer the agent. Following ARPO~\cite{dong2025agentic}, we incorporate both trajectory-level sampling and Information-Gain driven partial sampling during the rollout phase to cover a comprehensive sampling scope. Specifically, our proposed \textbf{Information-Gain Driven Diverse Branching} involves the following four kry steps:\\

\noindent \textit{(i) Rollout Initialization}: Given a global rollout budget of $M$, the LLM first generates $N$ trajectories via standard sampling based on the input question $q$. The remaining budget $M - N$ is reserved for adaptive partial sampling, triggered dynamically during the generation process.\\

\noindent \textit{(ii) Information-Gain Monitoring:} To determine pivotal branching points, we quantify the utility of the current observation $o_t$ using an Information-Gain (IG) Score. This is formally defined as the Pointwise Mutual Information (PMI) between the observation and the ground-truth answer $y^*$, conditioned on the context history $\mathcal{H}_t$:
\begin{equation}
\label{eq:ig_score}
\text{Score}_{\text{IG}}(o_t) = \log P(y^* \mid \mathcal{H}_t, o_t) - \log P(y^* \mid \mathcal{H}_t).
\end{equation}
This metric measures the reduction in uncertainty regarding the final answer $y^*$ provided by the current search result. \\ 

\noindent \textit{(iii) Dynamic Prompting Interventions:} Based on the IG score and search history, SIGHT dynamically triggers one of three exploration strategies by injecting specific hints into the context: \textbf{Redundancy Check (De-duplication).} If the current search query $\alpha_t$ is semantically identical to a query in the history $\mathcal{H}_{<t}$, we inject a De-duplication hint: \textit{"This search query has been used before. Please switch to a different keyword or perspective."} This constraint compels the agent to immediately diversify its search strategy; \textbf{Low IG Score (Noise Trap).} If $\text{Score}_{\text{IG}} < \delta_{\text{low}}$, it signals a potential dead-end or ``noise trap'' where the tool result provides little value. To prevent Tunnel Vision, we inject a Reflection Hint: \textit{"Analyze the gap between the current tool result and the final goal. What is missing? Generate a new search query targeting the missing information."} This triggers reflective exploration, encouraging the agent to backtrack and refine its reasoning; \textbf{High IG Score (Pivotal State).} If $\text{Score}_{\text{IG}} > \delta_{\text{high}}$, the state is identified as pivotal. We trigger \textbf{Adaptive Branching}: a new branch is spawned by copying the current rollout context (including the SES-distilled evidence $e_t$). In this new branch, we inject a Pivotal State Hint: \textit{"Critical information found. If the above evidence supports a direct answer, answer directly; otherwise, consider other aspects of this question."} This strategy encourages the agent to either converge on the answer immediately or pivot to remaining sub-questions without losing valuable context. The specific prompts for the three hint types are detailed in Appendix ~\ref{app:prompt_hint}. Guided by Information Gain, this mechanism dynamically steers the model toward reasoning paths that promise the most valuable insights, avoiding wasteful exploration of low-utility areas. Crucially, it also preserves necessary negative samples to maintain intra-group diversity.\\

\noindent \textit{(iv) Termination}: The iteration terminates upon meeting one of two criteria: On the one hand, once the cumulative count of generated branches $\hat{Z}$ exhausts the partial sampling budget $M - N$, we cease branching and proceed with linear generation until the final answer is derived; On the other hand, if all initial paths terminate before reaching the budget, we supplement with additional trajectory-level samples to exhaust the budget.

By leveraging this efficient rollout mechanism, SIGHT promotes Information-Gain-aware exploration, allowing LLMs to effectively discern high-utility step-level search behaviors. \\

\noindent \textit{\textbf{Complexity Analysis.}}
By leveraging shared prefixes, SIGHT not only eliminates redundant attention computations but also caches search results for common paths, thereby minimizing the aggregate number of search queries. This dual optimization significantly enhances efficiency, reducing the complexity for an $n$-token trajectory from the standard $O(n^2)$ baseline to an interval of $[O(n \log n), O(n^2)]$.\\

\noindent \textit{\textbf{Internalization via Masking.}} During training rollouts, Dynamic Prompting Interventions (the ``Hints'' described in Sec.~\ref{subsec:igdb}) are visible to the agent to guide exploration. However, during the policy update phase, these hints are masked. This compels $\pi_\theta$ to internalize the logic of de-duplication, reflection, and branching. Following Search-R1~\cite{jin2025search}, we mask raw \texttt{<results>} but compute loss on \texttt{<self-evidence>}, as optimizing on this self-generated synthesis enables the agent to effectively internalize and distill external knowledge.

\subsection{Hierarchical Reward Modeling}
\label{sec:reward}
Serving as the primary optimization objective to steer the policy model, we employ a rule-based reward function consisting of two key signals: Outcome-Based Reward $\mathcal{R}_{\text{Ans}}$, assessing the correctness of the final answer; and Self-Evidence Support (SES) Reward $\mathcal{R}_{\text{SES}}$, evaluating the utility of retrieved information.\\

\noindent \textbf{\textit{(i) Outcome-Based Reward}}
To measure answer correctness, we compute the F1-score between the predicted answer token set $\hat{y}$ and the ground truth $y^*$. We explicitly encourage retrieval by adding a small bonus $\beta=0.1$ if a search action occurs:
\begin{equation}
\label{eq:r_ans}
\mathcal{R}_{\text{Ans}} = \text{F1}(\hat{y}, y^*) + \beta \cdot \mathbb{I}(\exists \text{``</search>''} \wedge \text{F1} > 0).
\end{equation}

\noindent \textbf{\textit{(ii) Self-Evidence Support (SES) Reward}}
$\mathcal{R}_{\text{SES}}$ serves as a partial credit to reinforce the distillation of noise-free evidence even when the final answer is incorrect. We concatenate all content within \texttt{<self-evidence>} tags into a sequence $o_{\text{SES}}$. A reward $\lambda=0.2$ is assigned if the ground truth $y^*$ is contained within this evidence:
\begin{equation}
\label{eq:r_ses}
\mathcal{R}_{\text{SES}} = \lambda \cdot \mathbb{I}\bigl( y^* \subseteq_{\text{norm}} o_{\text{SES}} \bigr).
\end{equation}

\noindent \textbf{\textit{(iii) Hierarchical Reward Design}}
We propose a hierarchical reward mechanism. First, we apply a Format Constraint: if the response violates the required tag structure (e.g., missing \texttt{<think>}), we assign a penalty $\mathcal{R}_{\text{Format}} \in \{-1, -0.5\}$. Only when the format is valid ($\mathcal{R}_{\text{Format}} = 0$) do we compute the SES and outcome rewards. The total reward is prioritized as follows:
\begin{equation}
\label{eq:r_overall}
\mathcal{R}_{\text{Total}} = \mathcal{R}_{\text{Format}} + \begin{cases}
    \mathcal{R}_{\text{Ans}}, & \text{if } \mathcal{R}_{\text{Ans}} > 0 \quad (\text{\scriptsize Correct Answer}); \\
    \mathcal{R}_{\text{SES}}, & \text{if } \mathcal{R}_{\text{Ans}} = 0 \land \mathcal{R}_{\text{SES}} > 0 \; (\text{\scriptsize Evidence Found}); \\
    0, & \text{otherwise}.
\end{cases}
\end{equation}
This hierarchy prioritizes format compliance, followed by answer correctness and evidence retrieval.

\begin{table*}[t]
\centering
\caption{Main results on Single-Hop and Multi-Hop QA datasets across different model scales. We report Exact Match (EM) and Token Cost (TC). \textbf{Avg-S}, \textbf{Avg-M}, and \textbf{Avg-All} denote the average results on Single-Hop, Multi-Hop, and all datasets, respectively. The symbols $^\dagger$ and $^*$ denote in-domain and out-of-domain datasets, respectively. The best results are highlighted in \textbf{bold}, and the second-best results are \underline{underlined}.}
\label{tab:main_results_combined}
\resizebox{\textwidth}{!}{%
\begin{tabular}{lcccccccccccccccccccc}
\toprule
\multirow{3}{*}{\textbf{Methods}} & \multicolumn{8}{c}{\textbf{Single-Hop QA}} & \multicolumn{10}{c}{\textbf{Multi-Hop QA}} & \multicolumn{2}{c}{\multirow{2}{*}{\textbf{Avg-All}}} \\
\cmidrule(lr){2-9} \cmidrule(lr){10-19}
 & \multicolumn{2}{c}{NQ$^\dagger$} & \multicolumn{2}{c}{TriviaQA$^*$} & \multicolumn{2}{c}{PopQA$^*$} & \multicolumn{2}{c}{\textbf{Avg-S}} & \multicolumn{2}{c}{HotpotQA$^\dagger$} & \multicolumn{2}{c}{2Wiki$^*$} & \multicolumn{2}{c}{Musique$^*$} & \multicolumn{2}{c}{Bamboogle$^*$} & \multicolumn{2}{c}{\textbf{Avg-M}} & \multicolumn{2}{c}{} \\
\cmidrule(lr){2-3} \cmidrule(lr){4-5} \cmidrule(lr){6-7} \cmidrule(lr){8-9} \cmidrule(lr){10-11} \cmidrule(lr){12-13} \cmidrule(lr){14-15} \cmidrule(lr){16-17} \cmidrule(lr){18-19} \cmidrule(lr){20-21}
 & EM$\uparrow$ & TC$\downarrow$ & EM$\uparrow$ & TC$\downarrow$ & EM$\uparrow$ & TC$\downarrow$ & EM$\uparrow$ & TC$\downarrow$ & EM$\uparrow$ & TC$\downarrow$ & EM$\uparrow$ & TC$\downarrow$ & EM$\uparrow$ & TC$\downarrow$ & EM$\uparrow$ & TC$\downarrow$ & EM$\uparrow$ & TC$\downarrow$ & EM$\uparrow$ & TC$\downarrow$ \\
\midrule

\multicolumn{21}{c}{\textbf{Qwen2.5-3B-Instruct}} \\
\midrule

\multicolumn{21}{l}{\textit{w/o Retrieval}} \\ 
\hspace{1em}Direct Reasoning & 3.71 & 0.00 & 28.71 & 0.00 & 8.20 & 0.00 & 13.54 & 0.00 & 7.42 & 0.00 & 6.06 & 0.00 & 2.54 & 0.00 & 14.40 & 0.00 & 7.60 & 0.00 & 10.15 & 0.00 \\
\hspace{1em}COT & 5.08 & 0.00 & 27.54 & 0.00 & 7.81 & 0.00 & 13.48 & 0.00 & 4.10 & 0.00 & 2.15 & 0.00 & 3.13 & 0.00 & 16.00 & 0.00 & 6.34 & 0.00 & 9.40 & 0.00 \\
\hspace{1em}GRPO & 14.65 & 0.00 & 41.41 & 0.00 & 16.99 & 0.00 & 24.35 & 0.00 & 17.38 & 0.00 & 25.78 & 0.00 & 3.52 & 0.00 & 30.40 & 0.00 & 19.27 & 0.00 & 21.45 & 0.00 \\
\midrule

\multicolumn{21}{l}{\textit{w/ Retrieval}} \\ 
\hspace{1em}Vanilla RAG & 1.56 & 1.00 & 13.67 & 1.00 & 11.91 & 1.00 & 9.05 & 1.00 & 5.66 & 1.00 & 3.13 & 1.00 & 0.98 & 1.00 & 8.80 & 1.00 & 4.64 & 1.00 & 6.53 & 1.00 \\
\hspace{1em}ReAct & 0.20 & 1.89 & 5.86 & 2.35 & 2.93 & 2.87 & 3.00 & 2.37 & 2.54 & 3.12 & 1.37 & 3.85 & 1.56 & 3.60 & 2.72 & 4.00 & 2.37 & 3.32 & 2.64 & 2.91 \\
\hspace{1em}Search-o1 & 16.99 & 0.90 & 47.66 & 0.85 & 27.34 & 1.15 & 30.66 & 0.97 & 25.20 & 1.19 & 24.22 & 1.62 & 5.86 & 0.69 & 28.80 & 0.77 & 21.02 & 1.07 & 25.15 & 1.02 \\
\hspace{1em}Search-R1 & 25.00 & 1.69 & 50.00 & 1.68 & 38.09 & 1.56 & 37.70 & 1.64 & 35.74 & 2.31 & 33.20 & 2.76 & 10.94 & 2.95 & 33.60 & 2.31 & 28.37 & 2.58 & 32.37 & 2.18 \\
\hspace{1em}Tree-GRPO & 23.24 & 1.47 & 50.20 & 1.36 & 35.35 & 1.45 & 36.26 & 1.42 & 32.62 & 1.76 & 26.56 & 2.06 & 9.57 & 2.17 & 26.40 & 1.82 & 23.79 & 1.95 & 29.13 & 1.72 \\
\hspace{1em}ARPO & \underline{28.32} & 2.93 & \underline{56.25} & 2.96 & \underline{39.26} & 3.00 & \underline{41.28} & 2.96 & \underline{41.60} & 3.06 & \underline{34.77} & 3.37 & \textbf{17.38} & 3.31 & \underline{34.40} & 3.06 & \underline{32.04} & 3.20 & \underline{36.00} & 3.10 \\
\rowcolor{gray!20} \hspace{1em}\textbf{Ours} & \textbf{29.30} & \textbf{1.19} & \textbf{56.84} & \textbf{1.26} & \textbf{41.21} & \textbf{1.22} & \textbf{42.45} & \textbf{1.22} & \textbf{41.99} & \textbf{1.83} & \textbf{35.94} & \textbf{2.15} & \underline{15.82} & \textbf{2.34} & \textbf{40.00} & \textbf{1.95} & \textbf{33.44} & \textbf{2.07} & \textbf{37.30} & \textbf{1.71} \\

\midrule[\heavyrulewidth] 

\multicolumn{21}{c}{\textbf{Qwen2.5-7B-Instruct}} \\
\midrule

\multicolumn{21}{l}{\textit{w/o Retrieval}} \\ 
\hspace{1em}Direct Reasoning & 0.00 & 0.00 & 0.20 & 0.00 & 0.00 & 0.00 & 0.07 & 0.00 & 0.00 & 0.00 & 0.40 & 0.00 & 0.00 & 0.00 & 0.00 & 0.00 & 0.10 & 0.00 & 0.08 & 0.00 \\
\hspace{1em}COT & 0.00 & 0.00 & 0.60 & 0.00 & 0.00 & 0.00 & 0.20 & 0.00 & 0.00 & 0.00 & 0.20 & 0.00 & 0.00 & 0.00 & 0.00 & 0.00 & 0.05 & 0.00 & 0.11 & 0.00 \\
\hspace{1em}GRPO & 16.21 & 0.00 & 52.54 & 0.00 & 20.31 & 0.00 & 29.69 & 0.00 & 23.83 & 0.00 & 29.10 & 0.00 & 5.86 & 0.00 & 36.80 & 0.00 & 23.90 & 0.00 & 26.38 & 0.00 \\
\midrule

\multicolumn{21}{l}{\textit{w/ Retrieval}} \\ 
\hspace{1em}Vanilla RAG & 9.96 & 1.00 & 33.20 & 1.00 & 28.10 & 1.00 & 23.76 & 1.00 & 14.80 & 1.00 & 2.90 & 1.00 & 3.90 & 1.00 & 13.60 & 1.00 & 8.82 & 1.00 & 15.22 & 1.00 \\
\hspace{1em}ReAct & 8.59 & 1.45 & 35.16 & 1.46 & 22.66 & 1.45 & 22.14 & 1.45 & 18.95 & 1.91 & 15.63 & 2.16 & 8.98 & 2.29 & 28.80 & 1.79 & 18.09 & 2.04 & 19.82 & 1.79 \\
\hspace{1em}Search-o1 & 19.53 & 0.77 & 54.70 & 0.74 & 26.80 & 0.85 & 33.66 & 0.79 & 35.70 & 1.39 & 36.50 & 1.94 & 11.10 & 1.53 & 34.40 & 1.17 & 29.45 & 1.51 & 31.25 & 1.20 \\
\hspace{1em}Search-R1 & 25.98 & 1.57 & \underline{59.18} & 1.55 & 41.80 & 1.69 & 42.32 & 1.60 & 36.91 & 2.95 & 36.91 & 2.95 & 15.43 & 2.78 & \textbf{47.20} & 2.15 & 34.94 & 2.49 & 38.10 & 2.11 \\
\hspace{1em}Tree-GRPO & 30.86 & 1.26 & 57.81 & 1.24 & \textbf{44.14} & 1.27 & 44.27 & 1.26 & 40.43 & 1.70 & 32.23 & 2.09 & 13.87 & 2.27 & 39.20 & 1.86 & 31.43 & 1.98 & 36.93 & 1.67 \\
\hspace{1em}ARPO & \underline{31.45} & 1.93 & \textbf{59.20} & 1.91 & 42.60 & 1.95 & \underline{44.40} & 1.93 & \underline{46.10} & 2.18 & \underline{41.20} & 2.40 & \underline{18.60} & 2.50 & 39.20 & 2.18 & \underline{36.26} & 2.31 & \underline{39.75} & 2.15 \\
\rowcolor{gray!20} \hspace{1em}\textbf{Ours} & \textbf{31.45} & \textbf{1.10} & 59.00 & \textbf{1.10} & \underline{43.20} & \textbf{1.11} & \textbf{44.53} & \textbf{1.11} & \textbf{46.30} & \textbf{1.73} & \textbf{43.00} & \textbf{1.98} & \textbf{18.60} & \textbf{2.18} & \underline{43.20} & \textbf{1.92} & \textbf{37.75} & \textbf{1.95} & \textbf{40.66} & \textbf{1.59} \\

\bottomrule
\end{tabular}%
}
\end{table*}

\section{Experiments}

\noindent In this section, we conduct extensive experiments, aiming to answer the following Research Questions (RQs):
\begin{itemize}[leftmargin=*]
    \item \textbf{RQ1:} How does SIGHT enhance accuracy while reducing search costs compared to existing solutions?
    
    \item \textbf{RQ2:} What are the individual contributions of the core components to the SIGHT's overall effectiveness?
        
    \item \textbf{RQ3:} How do SIGHT's training dynamics evolve regarding efficiency, stability, and model scalability?
    
    \item \textbf{RQ4:} Can SIGHT maintain robust performance in complex, long-horizon ($\ge$ 3 hops) reasoning scenarios?
\end{itemize}

\subsection{Experimental Setup}

\textit{\textbf{Datasets.}} We evaluate our framework on seven diverse QA benchmarks, categorized into two groups. The \textbf{single-hop} datasets include Natural Questions (NQ)~\cite{kwiatkowski2019natural}, TriviaQA~\cite{joshi2017triviaqa}, and PopQA~\cite{mallen2022not}. The \textbf{multi-hop} datasets, which require complex reasoning, include HotpotQA~\cite{yang2018hotpotqa}, 2WikiMultihopQA (2Wiki)~\cite{ho2020constructing}, Musique~\cite{trivedi2022musique}, and Bamboogle~\cite{press2022measuring}. 
To address the issues of temporal ambiguity and unclear queries, we adopt the refined data processing strategy from DeepResearcher~\cite{zheng2025deepresearcher}. Specifically, we construct a training set of 18,000 samples extracted from their processed corpus, comprising 6,000 samples from NQ and 12,000 samples from HotpotQA. We intentionally allocate a higher proportion to HotpotQA to emphasize multi-hop reasoning scenarios, which better reflect the complex multi-turn search behaviors essential for in-depth research tasks. For evaluation, we directly employ the corresponding test sets provided by DeepResearcher~\cite{zheng2025deepresearcher}.\\

\noindent \textit{\textbf{Evaluation Metrics.}} We primarily employ \textbf{Exact Match (EM)} accuracy as the standard metric to measure the consistency between the predicted answer and the ground truth. Furthermore, to quantify the computational cost and search efficiency, we report the \textbf{Average Tool Calls (TC)}, which measures the frequency of search interactions required per query.\\

\noindent \textit{\textbf{Implementation Details.}} To simulate realistic open-domain search scenarios, we discard the original context documents provided in the QA datasets~\cite{yang2018hotpotqa,ho2020constructing,trivedi2022musique} and instead utilize the December 2018 Wikipedia dump~\cite{karpukhin2020dense} as our external knowledge source. We employ E5-base-v2~\cite{wang2022text} as the retrieval engine. By default, the retriever fetches the top-3 most relevant documents for each query. we run experiments using \textbf{Qwen2.5-3B-Instruct} and \textbf{Qwen2.5-7B-Instruct} as backbone models.  We train the models until convergence with a total of 250 steps. Regarding the specific hyperparameters for our SIGHT framework, we set the information-gain thresholds as $\delta_{\text{low}}=0$ and $\delta_{\text{high}}=0.5$. For the De-duplication mechanism, the F1 similarity threshold is set to 0.8 to effectively filter redundant reasoning paths. Full implementation details for all experiments are provided in Appendix ~\ref{app:imp_details}.\\

\noindent \textit{\textbf{Baselines.}} We compare SIGHT against two categories of methods to validate the effectiveness of our framework: 
(i) \textit{Generation w/o Retrieval}, which evaluates the model's internal knowledge and reasoning capabilities. This category includes direct generation (\textbf{Direct Reasoning}), Chain-of-Thought prompting (\textbf{CoT})~\cite{wei2022chain}, and the standard Group Relative Policy Optimization without retrieval (\textbf{GRPO})~\cite{shao2024deepseekmath}.
(ii) \textit{Generation w/ Retrieval}, which assesses the model's ability to leverage external information. This category covers: standard retrieval-augmented generation (\textbf{Vanilla RAG}); the interleaved reasoning and acting framework (\textbf{ReAct})~\cite{yao2022react}; the agentic search method \textbf{Search-o1}~\cite{li2025search}; and the agentic RL baseline \textbf{Search-R1}~\cite{jin2025search} alongside its variants \textbf{Tree-GRPO} \cite{ji2025tree} and \textbf{ARPO} \cite{dong2025agentic}, which utilize different branching exploration strategies. \\


\subsection{Main Results (RQ1)}
\label{sec:exp_main_results}

Table~\ref{tab:main_results_combined} presents the comparative results on both Single-Hop and Multi-Hop QA benchmarks across Qwen2.5-3B and Qwen2.5-7B scales. In a fair setting, SIGHT consistently outperforms state-of-the-art baselines, firmly establishing its superiority. Moreover, we highlight the following insights regarding accuracy and cost. \\

\noindent \textit{\textbf{Limitations of Internal Knowledge and Vanilla RL:} } Standard prompting fails on open-domain QA tasks due to the lack of access to external information ($<11\%$ EM on 3B, $<1\%$ on 7B). While GRPO significantly improves reasoning (21.45\%/26.38\% on 3B/7B), its performance remains inadequate compared to retrieval-augmented methods, confirming that reasoning improvements alone are insufficient without current external knowledge.\\

\noindent \textit{\textbf{Inconsistency of Static Retrieval:} }
Traditional methods (Vanilla RAG, ReAct) are limited by one-pass retrieval or myopic heuristics, failing to adapt to dynamic information needs. While Search-o1 performs well (25.15\% on 3B) via test-time compute, it lacks policy optimization, limiting adaptability. This confirms that static or heuristic search remains inadequate for multi-hop reasoning.\\

\noindent \textit{\textbf{Robust Performance and Efficiency of SIGHT:} }
\textit{(i) Overall Superiority:} On Qwen2.5-3B, SIGHT achieves the highest average EM of 37.30\%, surpassing ARPO (36.00\%) and Search-R1 (32.37\%). This advantage scales to the 7B model, where SIGHT reaches 40.66\%, outperforming ARPO (39.75\%); \textit{(ii) Advantage in Complex Reasoning:} SIGHT shows significant gains in Multi-Hop scenarios. On 3B, it outperforms the runner-up ARPO by substantial margins in Multi-Hop settings (33.44\% vs. 32.04\%), while maintaining robust Single-Hop performance (42.45\%). Notably, on the challenging Bamboogle dataset, SIGHT achieves a +5.6\% improvement over ARPO (40.00\% vs. 34.40\%), demonstrating superior handling of long-horizon dependencies; \textit{(iii) Search Efficiency:} SIGHT achieves these results with significantly lower computational and search costs. On the 3B model, SIGHT records an average Tool Call (TC) of 1.71, drastically lower than ARPO (3.10). Similarly, on the 7B model, SIGHT maintains superior efficiency (TC 1.59 vs. 2.15). This validates that compared to uncertainty-based branching methods, SIGHT's information-gain guidance directs exploration toward more correct solution spaces, thereby significantly reducing tool calls.

\subsection{Ablation Studies (RQ2)}
\label{sec:ablation}

To validate the contribution of each component in SIGHT, we conduct a comprehensive ablation study by removing key mechanisms individually and in combination. The summarized results are presented in Table~\ref{tab:ablation}. The detailed performance breakdown across all individual datasets is provided in Appendix~\ref{app:Ablation}.\\

\noindent \textit{\textbf{Superiority of Dynamic Branching Exploration.}}
We first evaluate the branching advantage. Setting (\textit{w/o whole IG Branching}) degrades the system to a standard linear generation process equipped only with the SES mechanism. The universal performance drop across both Single-Hop (41.54\%) and Multi-Hop (32.59\%) benchmarks compared to the Full Model (S-Hop: 42.45\%, M-Hop: 33.44\%) confirms that dynamic branching exploration is inherently superior to linear generation, as linear paths often fail to cover the search space adequately.\\

\noindent \textit{\textbf{Impact of SES Mechanism.}}
The ablation of the Self-Evidence Support (SES) reveals divergent trends across task types. In Single-Hop scenarios, retaining the SES tag without its optimization reward (\textit{w/o} $\mathcal{R}_{\text{SES}}$, Setting 3) results in a notable performance drop (40.75\%) compared to completely removing the SES module (42.32\%). This suggests that for simple queries, unaligned evidence generation acts as a distraction. Conversely, in Multi-Hop scenarios, completely removing the SES module (\textit{w/o whole SES}, Setting 4) leads to a clear decline (32.82\%), highlighting that active noise filtration is essential for complex reasoning chains.\\

\noindent \textit{\textbf{Role of Dynamic Prompting Interventions in Complex Reasoning.}}
We observe a significant disparity regarding the impact of explicit interventions. In Single-Hop tasks, the setting (\textit{w/o Dyn. Interventions}) outperforms the setting (\textit{w/o SES \& Dyn. Interv.}) (40.88\% vs. 39.97\%), indicating that the SES Mechanism alone confers benefits for simple retrieval even without explicit hints.
However, a crucial reversal emerges in Multi-Hop scenarios. The performance of (\textit{w/o Dyn. Interventions}) suffers the most severe drop to 31.75\%, falling even below the setting (\textit{w/o SES \& Dyn. Interv.}, 32.75\%). This counter-intuitive finding suggests that retaining complex structures (Branching + SES) without prompt direction increases cognitive load. The agent struggles to utilize filtered evidence effectively without directive prompts like ``Reflection'' or ``De-duplication,'' leading to suboptimal search behaviors. Thus, Dynamic Prompting Interventions bridge the gap, enabling the model to leverage structural complexity for robust multi-hop reasoning. Furthermore, the performance drop in (\textit{w/o Dyn. Interventions}) ($\Delta 1.69\%$) is notably larger than that in (\textit{w/o whole IG Branching}) ($\Delta 0.85\%$). This underscores that within the IG Branching framework, the guidance provided by hint tokens is paramount; without them, the sophisticated branching architecture cannot be effectively utilized.

\begin{table}[t]
    \centering
    \caption{Ablation study on different SIGHT components. The \textit{w/o} denotes removing specific components and ``whole'' indicates removing the entire module. We report Average Exact Match (EM) and Toll Call (TC) across Single-Hop (S-Hop), Multi-Hop (M-Hop), and all datasets. }
    \label{tab:ablation}
    \resizebox{\linewidth}{!}{
    \begin{tabular}{lcccccc}
        \toprule
        \multirow{2}{*}{\textbf{Method}} & \multicolumn{2}{c}{\textbf{S-Hop}} & \multicolumn{2}{c}{\textbf{M-Hop}} & \multicolumn{2}{c}{\textbf{Avg-All}} \\
        \cmidrule(lr){2-3} \cmidrule(lr){4-5} \cmidrule(lr){6-7}
        & EM$\uparrow$ & TC$\downarrow$ & EM$\uparrow$ & TC$\downarrow$ & EM$\uparrow$ & TC$\downarrow$ \\
        \midrule
        
        \rowcolor{gray!20} SIGHT (Full) & 42.45 & 1.22 & 33.44 & 2.07 & 37.30 & 1.71 \\
        \midrule
        
        \multicolumn{7}{l}{\textit{Impact of Exploration Structure}} \\
        \hspace{1em}\textit{w/o} whole IG Branching & 41.54 & 1.86 & 32.59 & 2.22 & 36.42 & 2.07 \\
        \midrule
        
        \multicolumn{7}{l}{\textit{Impact of SES Mechanism}} \\
        \hspace{1em}\textit{w/o} $\mathcal{R}_{\text{SES}}$ (Reward Only) & 40.75 & 1.06 & 33.23 & 1.83 & 36.45 & 1.50 \\
        \hspace{1em}\textit{w/o} whole SES Mechanism & 42.32 & 1.17 & 32.82 & 2.03 & 36.89 & 1.66 \\
        \midrule
        
        \multicolumn{7}{l}{\textit{Impact of Interventions}} \\
        \hspace{1em}\textit{w/o} Dyn. Interventions & 40.88 & 2.03 & 31.75 & 2.05 & 35.66 & 2.03 \\
        \hspace{1em}\textit{w/o} SES \& Dyn. Interv. & 39.97 & 1.65 & 32.75 & 1.97 & 35.85 & 1.65 \\
        \bottomrule
    \end{tabular}
    }
\end{table}

\begin{figure}
    \centering
    \includegraphics[width=1\linewidth]{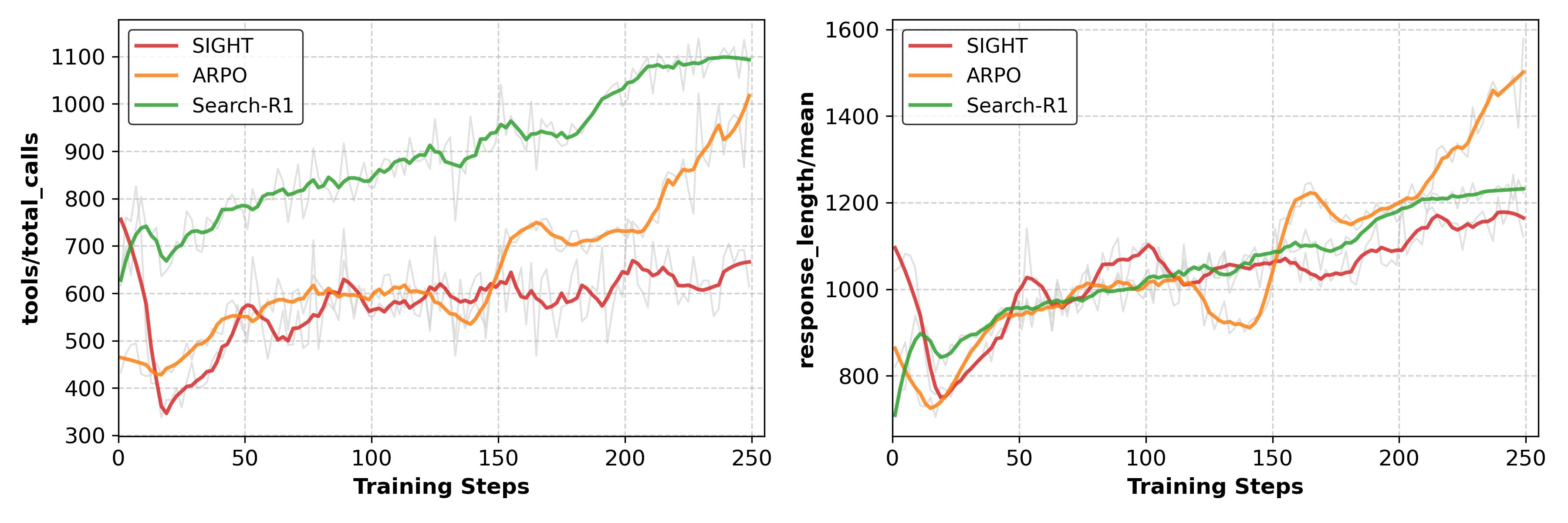}
    \caption{Training dynamics comparison among Search-R1, ARPO and our proposed SIGHT, where we report the Tool Calls and Response Length of trained models.}
    \label{fig:training_dynamics}
\end{figure}

\begin{figure*}[t!]
    \centering
    \includegraphics[width=1\linewidth]{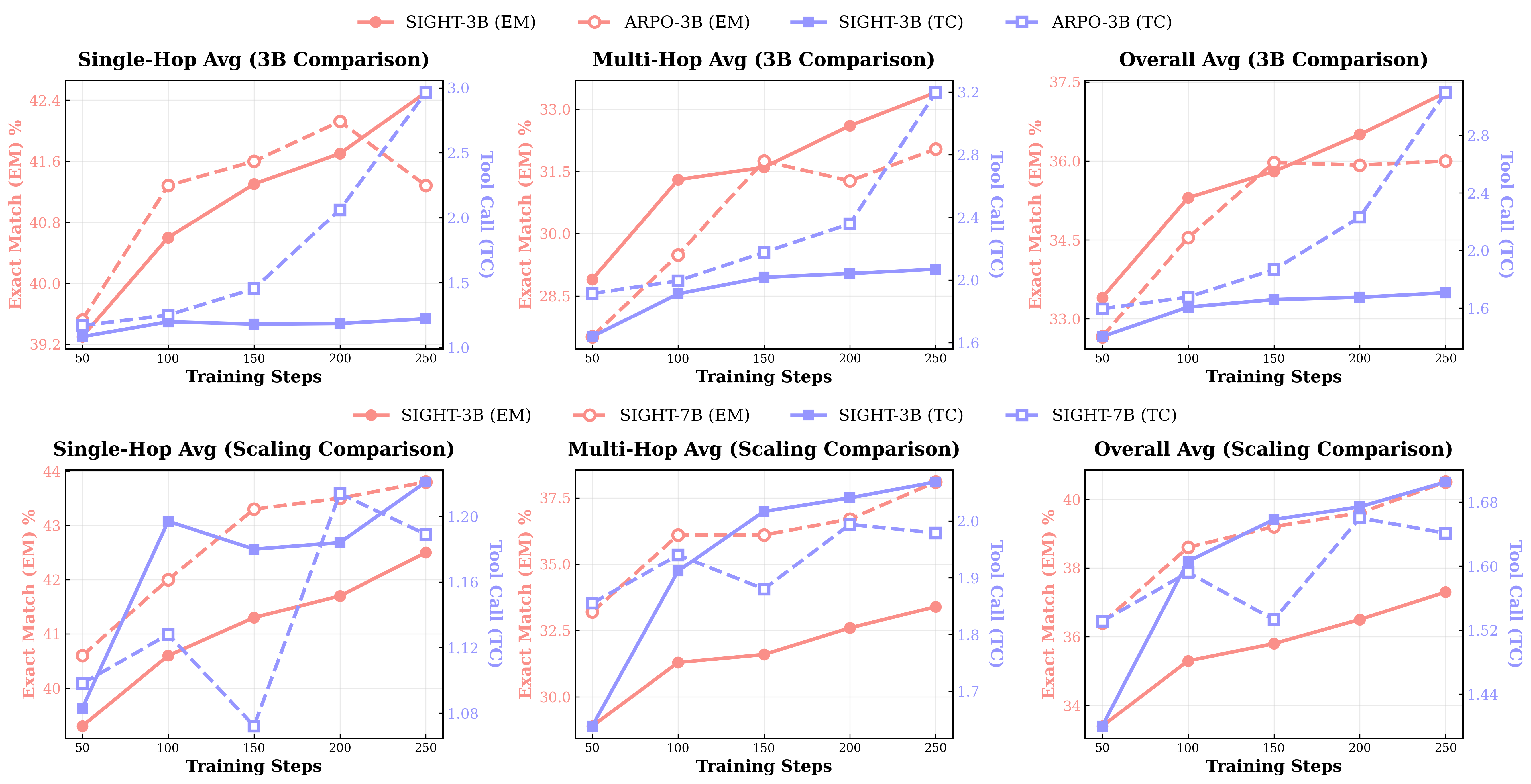}
    \caption{Training evolution analysis of Reasoning Performance (EM) and Search Cost (TC) across Qwen2.5-3B-Instruct and Qwen2.5-7B-Instruct on evaluation datasets including Single-Hop and Multi-Hop questions.}
    \label{fig:val_test}
\end{figure*}

\subsection{Training Dynamics Analysis (RQ3)}

To verify the efficiency and stability of SIGHT, we analyze the training dynamics from two perspectives: computational overhead during training and the evolution of model performance.\\

\noindent \textit{\textbf{Training Efficiency and Resource Consumption.} }
Figure \ref{fig:training_dynamics} illustrates the trajectory of total tool calls (TC) and response length throughout the training process. Compared to Search-R1, which relies on linear generation without shared prefixes, both branching-based methods (SIGHT and ARPO) significantly reduce the total TC overhead by reusing common reasoning paths. 
Notably, as training progresses into later stages, SIGHT achieves a lower total TC compared to the entropy-based ARPO. This validates the superiority of our Information-Gain (IG) Driven branching: unlike heuristic metrics that explore ambiguous but irrelevant paths, IG guidance precisely navigates the search space, focusing exploration solely on high-utility branches for greater resource efficiency.\\

\noindent \textbf{\textit{Evolution of Reasoning Capabilities.}} Figure \ref{fig:val_test} tracks the Average Exact Match (EM) and inference-time Tool Calls (TC) across seven datasets (categorized into Single-Hop and Multi-Hop) at 50-step intervals. Regarding the comparison between SIGHT and ARPO (3B), as shown in the top row of Figure \ref{fig:val_test}, SIGHT consistently outperforms ARPO in Multi-Hop scenarios throughout the training, while its Single-Hop capabilities steadily improve. Crucially, SIGHT achieves this superior accuracy while maintaining a stable and low tool call frequency, standing in stark contrast to ARPO's escalating computational costs during training. Furthermore, regarding the scalability analysis (3B vs. 7B), the bottom row of Figure \ref{fig:val_test} confirms SIGHT's scalability, with the 7B model significantly outperforming the 3B version across all tasks. This suggests that stronger intrinsic reasoning capabilities amplify SIGHT's effectiveness in navigating complex solution spaces.


\subsection{Complex Multi-Hop Analysis (RQ4)}
\label{subsec:rq4-3hop}

To rigorously assess the robustness of SIGHT in long-horizon reasoning scenarios, we conduct a specialized evaluation on a curated subset of questions requiring \textbf{at least 3 hops} of reasoning (details in Appendix~\ref{app:Multi-Hop}). These samples represent the most challenging "noise traps" where error accumulation typically leads to reasoning collapse. As presented in Table~\ref{tab:rq4-3hop}, SIGHT demonstrates exceptional resilience, achieving an Exact Match (EM) score of \textbf{31.98\%}. This significantly outperforms the strong Agentic RL baseline ARPO (30.46\%) and Search-R1 (25.36\%) by margins of 1.52\% and 6.62\%, respectively. \textbf{Efficiency vs. Accuracy Trade-off:} Even in such complex multi-hop scenarios, SIGHT consistently maintains its efficiency advantage. Remarkably, it achieves this superior performance with significantly fewer average tool calls (\textbf{2.21}) compared to ARPO (2.71) and Search-R1 (3.42). This inverse correlation between step count and accuracy provides empirical evidence for the "Tunnel Vision" hypothesis: while baselines like Search-R1 struggle with redundant queries and get trapped in noisy loops (inflating the step count), SIGHT's Information-Gain guidance effectively \textit{navigates} the reasoning space with surgical precision. By prioritizing high-utility paths and implicitly avoiding low-value explorations, SIGHT consistently converges to the correct answer more efficiently, proving to be both robust and computationally economical for deep reasoning tasks.

\begin{table}
\setlength{\abovecaptionskip}{0cm}   
\setlength{\belowcaptionskip}{0cm}   
\centering
\caption{Exact Match (EM) and Tool Calls (TC) on 3-hop and above multi-hop questions.}
\label{tab:rq4-3hop}

\newcommand{\base}{_{\Delta_{\text{base}}}} 
\newcommand{\imp}[1]{_{\textcolor{teal}{#1}}} 

\begin{tabular*}{0.95\columnwidth}{@{\extracolsep{\fill}}lcc}
\toprule
\textbf{Method} & \textbf{EM (\%)} & \textbf{TC} \\
\midrule
Search-R1 & $25.36\base$ & $3.42\base$ \\
ARPO & $30.46\imp{+5.10}$ & $2.71\imp{-0.71}$ \\
\textbf{SIGHT (Ours)} & $\mathbf{31.98}\imp{+6.62}$ & $\mathbf{2.21}\imp{-1.21}$ \\
\bottomrule
\end{tabular*}
\vspace{-3mm}
\end{table}

\section{Conclusion}
In this work, we present SIGHT, a noise-aware Agentic RL framework for multi-turn search tasks, designed to guide LLMs in exploring step-wise search behaviors under Information-Gain guidance. SIGHT enhances the standard agentic RL paradigm by integrating Self-Evidence Support (SES) for active noise filtration and Information-Gain Driven Branching Exploration. Extensive experiments demonstrate that SIGHT significantly outperforms advanced baselines in single-hop and multi-hop QA tasks, achieving superior accuracy with reduced search costs. Crucially, our IG guidance precisely navigates the search space, focusing exploration solely on high-utility branches for greater resource efficiency, making SIGHT a promising solution for long-horizon, multi-turn tasks.

\bibliographystyle{ACM-Reference-Format}
\bibliography{sample-base}

\appendix
\newpage
\section{Experimental details}
\label{app:exp_details}  

\subsection{Implementation Details}
\label{app:imp_details}

We employ Qwen2.5-3B-Instruct and Qwen2.5-7B-Instruct as our backbone models. The experiments for the 3B model are conducted on 4 NVIDIA H800 GPUs, while the 7B model experiments utilize 8 NVIDIA H800 GPUs. The training process uses the AdamW optimizer with a constant learning rate of $1\text{e-}6$.

Our standard training setup spans a total of 250 steps. We set the total training batch size to 128 and the PPO mini-batch size to 32. To ensure training stability, the KL divergence coefficient is set to 0.
During rollout generation, the maximum response length is capped at 4096 tokens, and the maximum number of tool calls is limited to 6. We configure the rollout parameters with a global rollout size ($N$) of 16, an initial sampling size of 8, and a beam size of 2.

Regarding the specific hyperparameters for our SIGHT framework, we set the information-gain thresholds as $\delta_{\text{low}}=0$ and $\delta_{\text{high}}=0.5$. For the De-duplication mechanism, the F1 similarity threshold is set to 0.8 to effectively filter redundant reasoning paths. Finally, the implementations of all baseline methods follow their original papers and official codebases.




\subsection{Ablation Study}
\label{app:Ablation}

The detailed performance breakdown of our ablation study across all individual datasets is provided in Table~\ref{tab:detailed_ablation}.

\begin{table*}[t]
\centering
\caption{\textbf{Detailed Ablation Results.} Full breakdown of Exact Match (EM) and Token Cost (TC) across all Single-Hop and Multi-Hop datasets. \textbf{Avg-S}, \textbf{Avg-M}, and \textbf{Avg-All} denote the average results on Single-Hop, Multi-Hop, and all datasets, respectively. The best model (SIGHT) is highlighted with a gray background.}
\label{tab:detailed_ablation}
\resizebox{\textwidth}{!}{%
\begin{tabular}{lcccccccccccccccccccc}
\toprule
\multirow{3}{*}{\textbf{Method}} & \multicolumn{8}{c}{\textbf{Single-Hop QA}} & \multicolumn{10}{c}{\textbf{Multi-Hop QA}} & \multicolumn{2}{c}{\multirow{2}{*}{\textbf{Avg-All}}} \\
\cmidrule(lr){2-9} \cmidrule(lr){10-19}
 & \multicolumn{2}{c}{NQ} & \multicolumn{2}{c}{TriviaQA} & \multicolumn{2}{c}{PopQA} & \multicolumn{2}{c}{\textbf{Avg-S}} & \multicolumn{2}{c}{HotpotQA} & \multicolumn{2}{c}{2Wiki} & \multicolumn{2}{c}{Musique} & \multicolumn{2}{c}{Bamboogle} & \multicolumn{2}{c}{\textbf{Avg-M}} & \multicolumn{2}{c}{} \\
\cmidrule(lr){2-3} \cmidrule(lr){4-5} \cmidrule(lr){6-7} \cmidrule(lr){8-9} \cmidrule(lr){10-11} \cmidrule(lr){12-13} \cmidrule(lr){14-15} \cmidrule(lr){16-17} \cmidrule(lr){18-19} \cmidrule(lr){20-21}
 & EM$\uparrow$ & TC$\downarrow$ & EM$\uparrow$ & TC$\downarrow$ & EM$\uparrow$ & TC$\downarrow$ & EM$\uparrow$ & TC$\downarrow$ & EM$\uparrow$ & TC$\downarrow$ & EM$\uparrow$ & TC$\downarrow$ & EM$\uparrow$ & TC$\downarrow$ & EM$\uparrow$ & TC$\downarrow$ & EM$\uparrow$ & TC$\downarrow$ & EM$\uparrow$ & TC$\downarrow$ \\
\midrule

\rowcolor{gray!20} SIGHT (Full) & 29.30 & 1.19 & 56.84 & 1.26 & 41.21 & 1.22 & 42.45 & 1.22 & 41.99 & 1.83 & 35.94 & 2.15 & 15.82 & 2.34 & 40.00 & 1.95 & 33.44 & 2.07 & 37.30 & 1.71 \\
\midrule

\multicolumn{21}{l}{\textit{Impact of Exploration Structure}} \\
\hspace{1em}\textit{w/o} whole IG Branching & 27.34 & 1.88 & 58.01 & 1.89 & 39.26 & 1.81 & 41.54 & 1.86 & 41.02 & 2.07 & 36.33 & 2.21 & 16.99 & 2.46 & 36.00 & 2.14 & 32.59 & 2.22 & 36.42 & 2.07 \\
\midrule

\multicolumn{21}{l}{\textit{Impact of SES Mechanism}} \\
\hspace{1em}\textit{w/o} $\mathcal{R}_{\text{SES}}$ (Reward Only) & 27.34 & 1.04 & 56.05 & 1.08 & 38.87 & 1.06 & 40.75 & 1.06 & 42.58 & 1.59 & 38.28 & 1.90 & 14.45 & 2.04 & 37.60 & 1.79 & 33.23 & 1.83 & 36.45 & 1.50 \\
\hspace{1em}\textit{w/o} whole SES Mechanism & 28.52 & 1.13 & 58.20 & 1.15 & 40.23 & 1.23 & 42.32 & 1.17 & 41.41 & 1.70 & 38.28 & 2.14 & 16.41 & 2.38 & 35.20 & 1.91 & 32.82 & 2.03 & 36.89 & 1.66 \\
\midrule

\multicolumn{21}{l}{\textit{Impact of Guidance}} \\
\hspace{1em}\textit{w/o} Dyn. Interventions & 26.76 & 1.99 & 55.66 & 2.00 & 40.23 & 2.00 & 40.88 & 2.00 & 39.84 & 2.01 & 36.91 & 2.06 & 15.04 & 2.13 & 35.20 & 2.01 & 31.75 & 2.05 & 35.66 & 2.03 \\
\hspace{1em}\textit{w/o} SES \& Dyn. Interv. & 27.34 & 1.21 & 53.52 & 1.21 & 39.06 & 1.22 & 39.97 & 1.21 & 38.67 & 1.75 & 36.33 & 1.99 & 16.02 & 2.25 & 40.00 & 1.89 & 32.75 & 1.97 & 35.85 & 1.65 \\
\bottomrule
\end{tabular}%
}
\end{table*}

\begin{figure}
    \centering
    \includegraphics[width=1\linewidth]{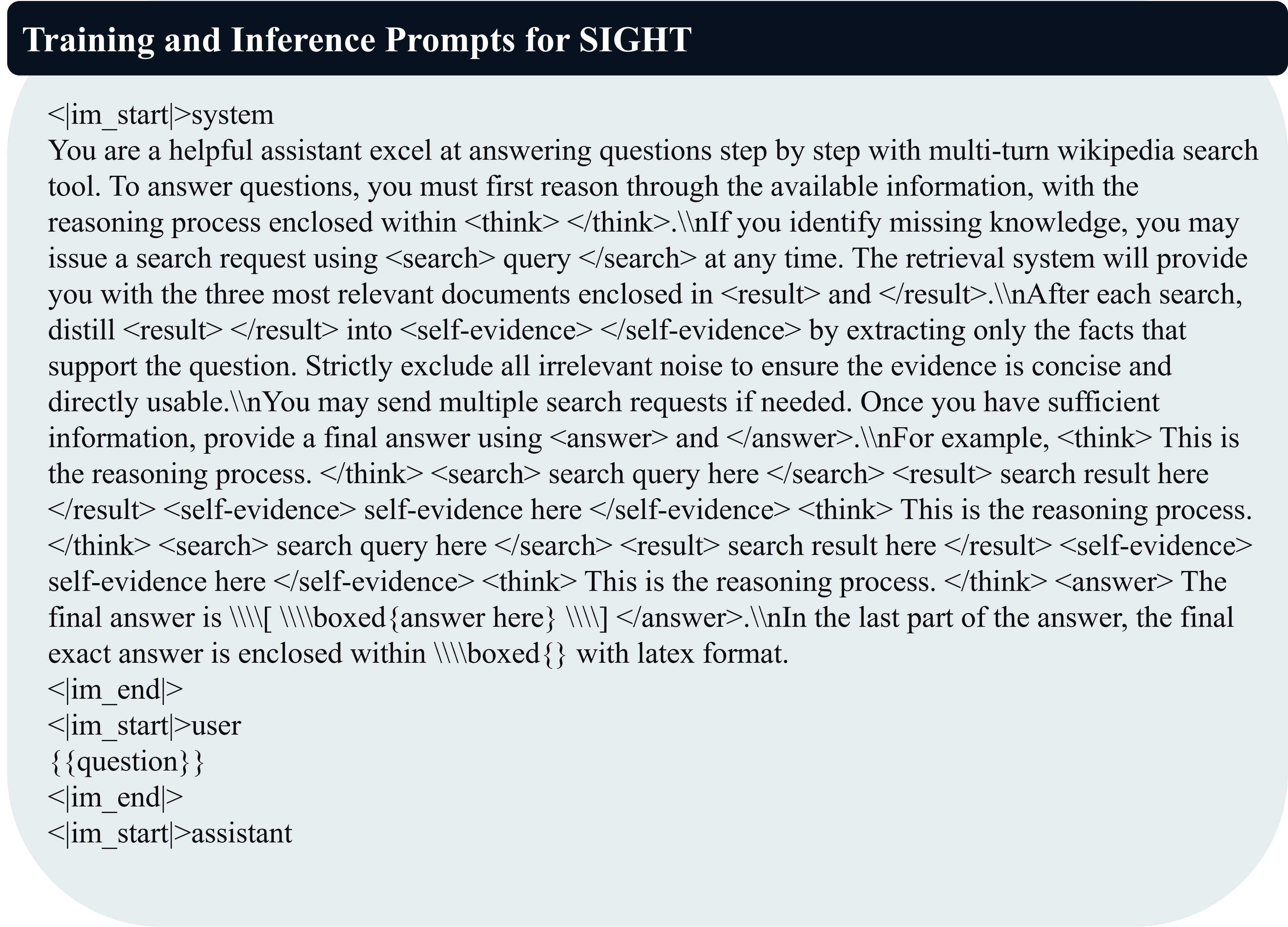}
    \caption{Training and Inference Prompts for SIGHT}
    \label{fig:prompt}
\end{figure}

\begin{figure}
    \centering
    \includegraphics[width=1\linewidth]{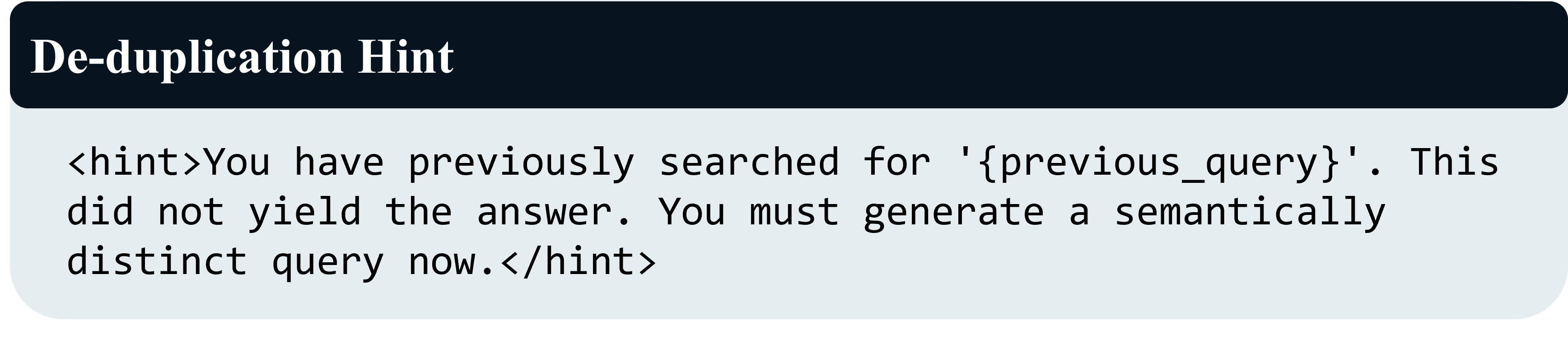}
    \caption{De-duplication Hint}
    \label{fig:hint1}
\end{figure}

\begin{figure}
    \centering
    \includegraphics[width=1\linewidth]{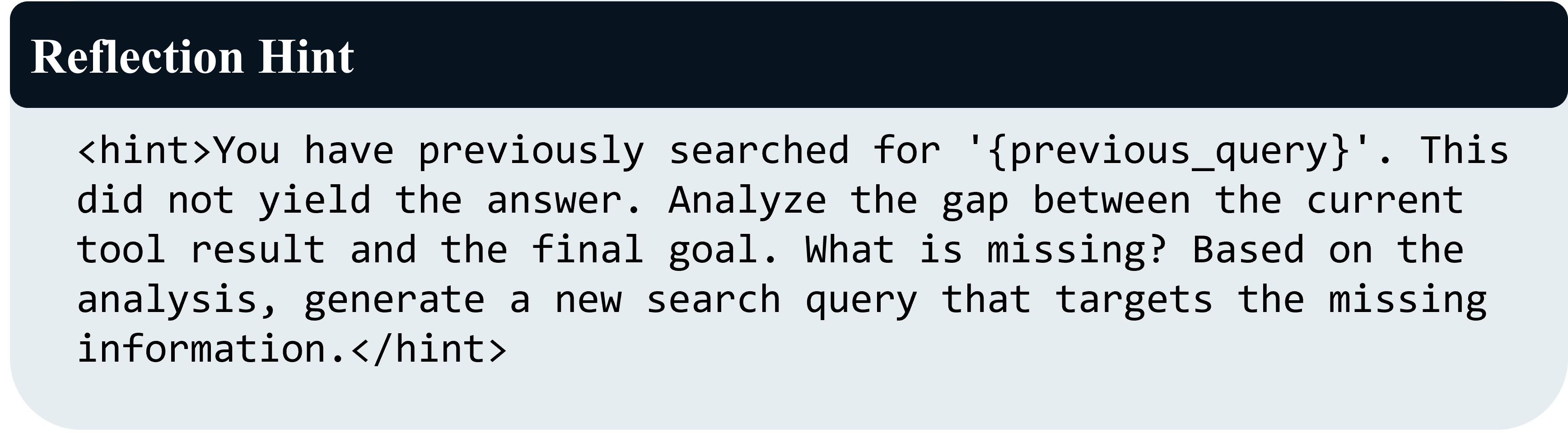}
    \caption{Reflection Hint}
    \label{fig:hint2}
\end{figure}

\begin{figure}
    \centering
    \includegraphics[width=1\linewidth]{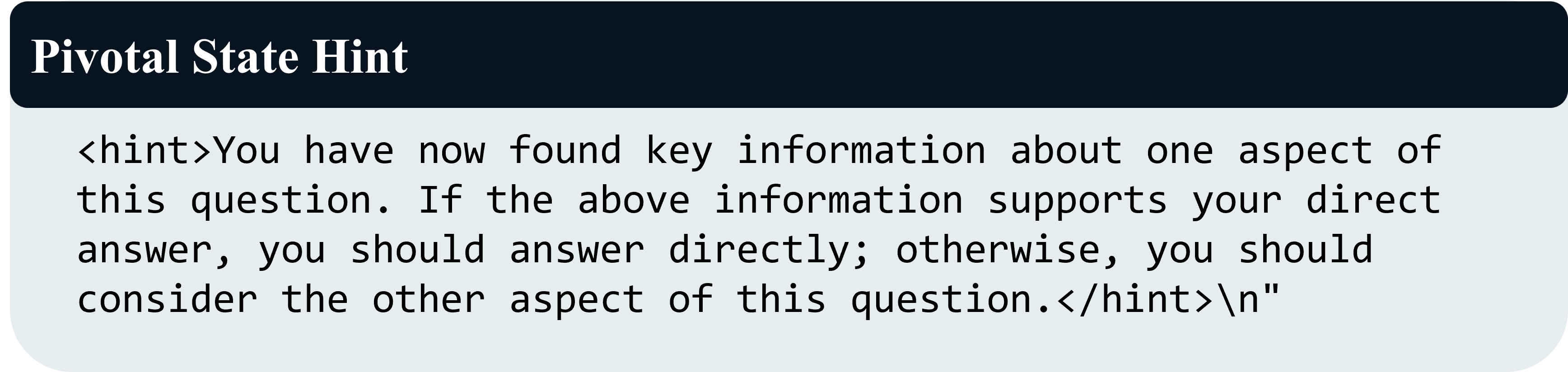}
    \caption{Pivotal State Hint}
    \label{fig:hint3}
\end{figure}

\subsection{Prompt and Hint}
\label{app:prompt_hint}

Figure \ref{fig:prompt} illustrates the overall system instruction for SIGHT, while Figures \ref{fig:hint1},\ref{fig:hint2} and \ref{fig:hint3} display the specific prompt templates for the \textbf{De-duplication Hint}, the \textbf{Reflection Hint}, and the \textbf{Pivotal State Hint}, respectively.

\subsection{Complex Multi-Hop Experiment}
\label{app:Multi-Hop}

To evaluate model performance on complex multi-hop reasoning, we constructed a curated subset from the original test sets. Specifically, we utilized \textbf{GPT-5.2} to identify questions requiring \textbf{at least 3 hops} of reasoning. This resulted in a total of 809 samples, distributed as follows: Musique (401), HotpotQA (204), 2Wiki (161), and Bamboogle (43).

\subsection{Training Dynamics Analysis }
\label{app:Dynamics}

\begin{figure}
    \centering
    \includegraphics[width=1\linewidth]{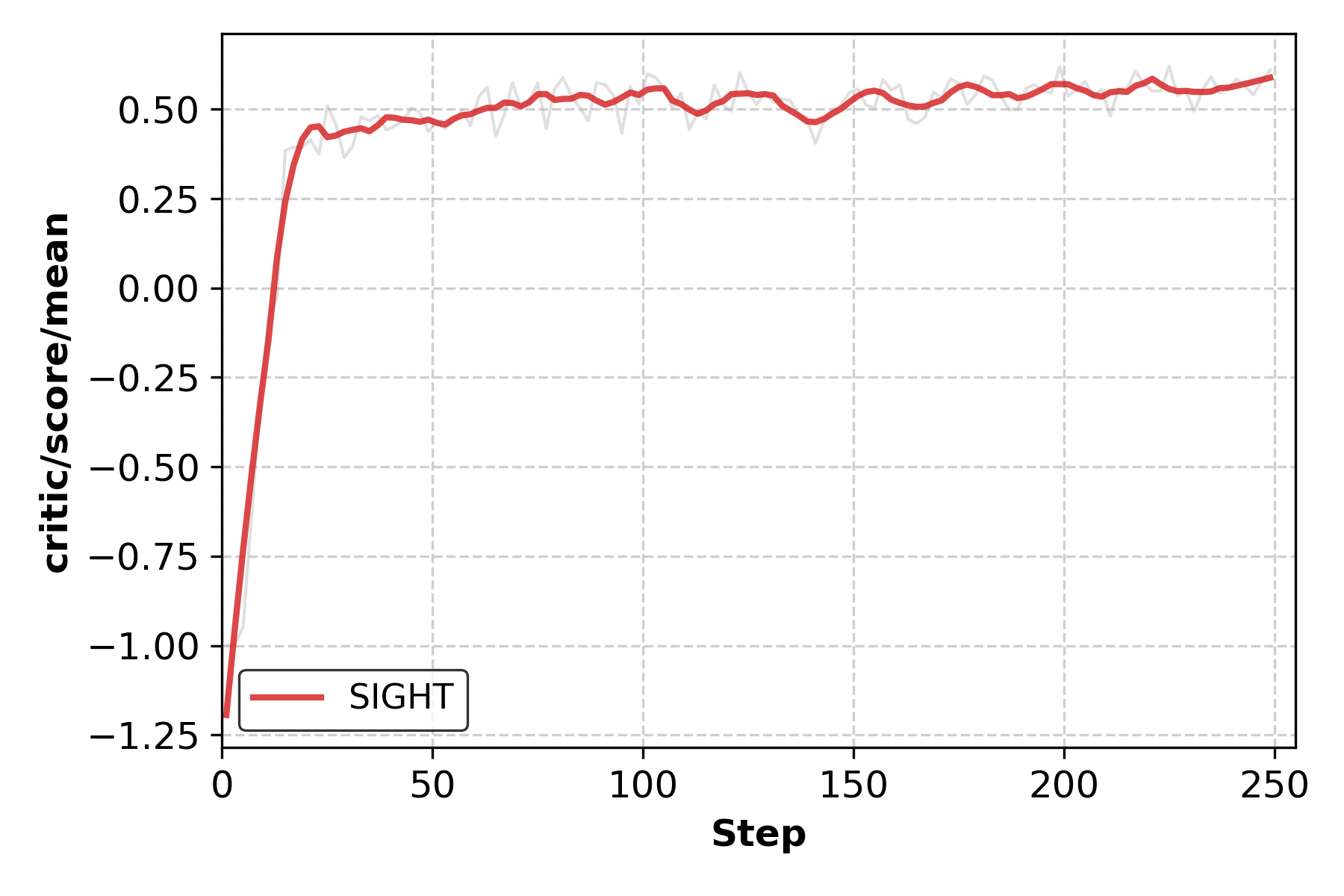}
    \caption{Training Reward}
    \label{fig:critic_score_mean_1}
\end{figure}

Figure~\ref{fig:critic_score_mean_1} illustrates the evolution of reward scores during the training process of SIGHT.

\begin{figure*}
    \centering
    \includegraphics[width=1\linewidth]{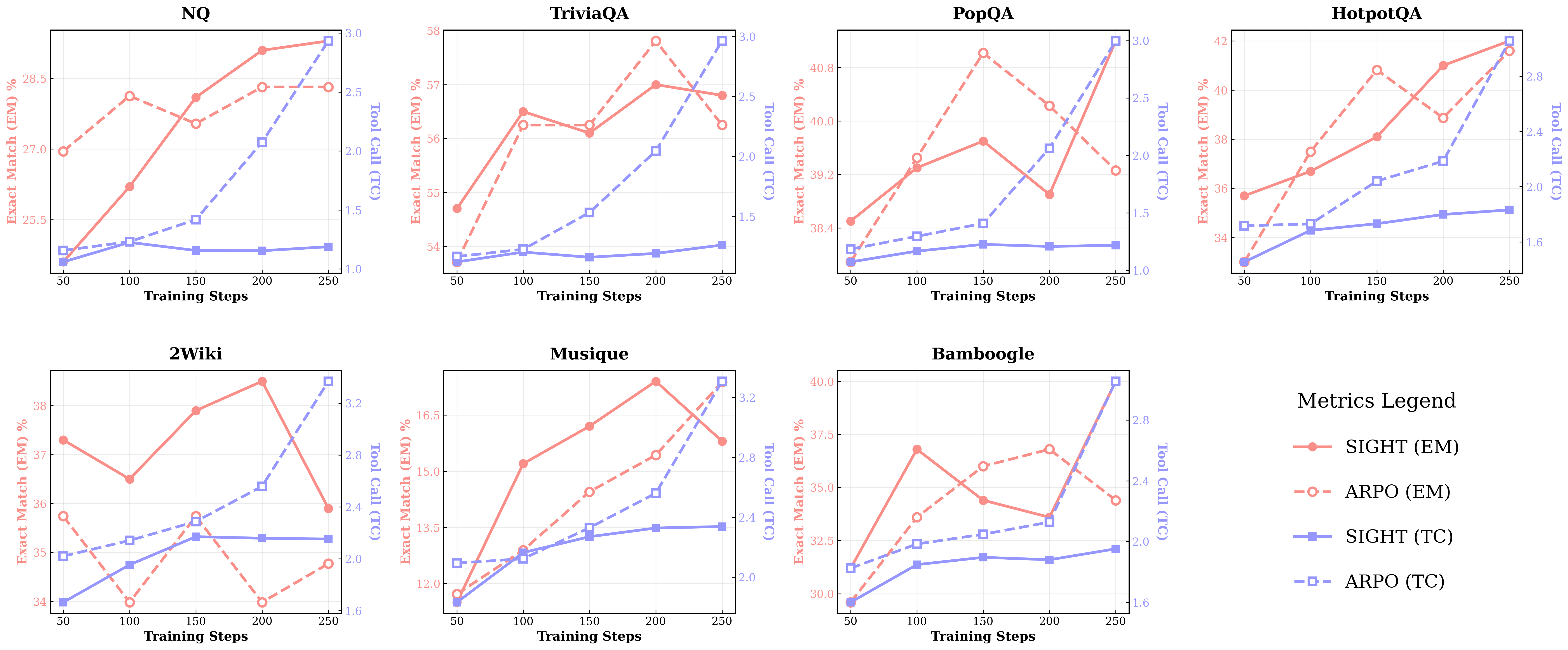}
    \caption{Training dynamics of SIGHT (solid) vs. ARPO (dashed) on the 3B model across seven datasets, showing Exact Match (red) and average Tool Calls (purple).}
    \label{fig:training_dynamics_3b_7datasets}
\end{figure*}

Figure~\ref{fig:training_dynamics_3b_7datasets} illustrates the detailed training dynamics across all seven datasets, demonstrating SIGHT's consistent superiority in both accuracy and search efficiency over ARPO.

\subsection{Case Study}
\label{app:Case}

To provide a concrete understanding of how SIGHT operates in complex reasoning scenarios, we present detailed case studies in Tables~\ref{tab:case_study_2}, \ref{tab:case_study_3}, and \ref{tab:triple_hint_case}.

\begin{table*}[h]
\centering
\caption{Comparison on a complex multi-hop query. \textbf{Baseline} fails to isolate the target entity, leading to a hallucinated answer. \textbf{SIGHT} successfully identifies the intermediate entity (\textit{Patricia Arquette}) triggered by the \textbf{Reflection Hint}, and then correctly retrieves her debut year guided by the \textbf{Information-Gain Hint}.}
\label{tab:case_study_2}
\begin{tabularx}{\linewidth}{lX}
\toprule
\multicolumn{2}{l}{\textbf{Question:} Find the actress who starred alongside Sigourney Weaver... and then determine her film debut.} \\
\multicolumn{2}{l}{\textbf{Ground Truth:} \textcolor{correctgreen}{Patricia Arquette (1987)}} \\
\midrule

\textbf{SIGHT} & \textcolor{taggray}{<think>} ...we need to first identify the film they were in together. \textcolor{taggray}{</think>} \\
\textbf{(Correct)} & \textcolor{taggray}{<search>} Sigourney Weaver... film debut \textcolor{taggray}{</search>} \\
 & \textcolor{taggray}{<result>} [Doc 2] ...\textbf{Holes} (2003)... stars Sigourney Weaver... \textcolor{evidenceblue}{\textbf{Patricia Arquette}}... \textcolor{taggray}{</result>} \\
 & \textcolor{hintorange}{<hint> You have previously searched... This did not yield the answer. Analyze the gap... generate a new search query that targets the missing information. </hint>} \\
 & \textcolor{taggray}{<think>} ...film ``Holes'' is the one... Now we need to find the actress who starred alongside... \textcolor{taggray}{</think>} \\
 & \textcolor{taggray}{<search>} actress who starred alongside... in the film ``Holes'' \textcolor{taggray}{</search>} \\
 & \textcolor{taggray}{<result>} ...Holes... stars... \textcolor{evidenceblue}{\textbf{Patricia Arquette}}... \textcolor{taggray}{</result>} \\
 & \textcolor{evidenceblue}{<self-evidence>} The actress who starred alongside... is Patricia Arquette. \textcolor{evidenceblue}{</self-evidence>} \\
 & \textcolor{hintorange}{<hint> You have now found key information... If the above information supports your direct answer, you should answer directly; otherwise, you should consider the other aspect... </hint>} \\
 & \textcolor{taggray}{<think>} Now that we know the actress is Patricia Arquette, we can find her film debut. \textcolor{taggray}{</think>} \\
 & \textcolor{taggray}{<search>} Patricia Arquette film debut \textcolor{taggray}{</search>} \\
 & \textcolor{taggray}{<result>} [Doc 2] ...made her feature film debut as Kristen Parker in `` '' (\textcolor{evidenceblue}{\textbf{1987}})... \textcolor{taggray}{</result>} \\
 & \textcolor{taggray}{<answer>} \textcolor{correctgreen}{\boxed{1987}} \textcolor{taggray}{</answer>} \\

\midrule

\textbf{Baseline} & \textcolor{taggray}{<think>} ...search for movies they have acted in together. \textcolor{taggray}{</think>} \\
\textbf{(Incorrect)} & \textcolor{taggray}{<search>} Sigourney Weaver... movie debut \textcolor{taggray}{</search>} \\
 & \textcolor{taggray}{<result>} [Doc 2] ...\textbf{Holes} (2003)... stars... Patricia Arquette... \textcolor{taggray}{</result>} \\
 & \textcolor{taggray}{<think>} ...film ``Holes'' is the one... Now, we need to find out the actress's film debut. \textcolor{taggray}{</think>} \\
 & \textcolor{taggray}{<search>} Sigourney Weaver... Holes film debut \textcolor{taggray}{</search>} \\
 & \textcolor{taggray}{<result>} ...[Irrelevant info about Andrew Davis, Warren Beatty]... \textcolor{taggray}{</result>} \\
 & \textcolor{wrongred}{<self-evidence> ...Sigourney Weaver's film debut is in 1986 with the movie ``Insurgent Territory''. </self-evidence>} \\
 & \textcolor{taggray}{<answer>} \textcolor{wrongred}{\boxed{1986}} \textcolor{taggray}{</answer>} \\

\bottomrule
\end{tabularx}
\end{table*}

\begin{table*}[h]
\centering
\caption{Case study on fine-grained information extraction. The \textbf{Baseline} correctly identifies multiple metrics but fails to distinguish the specific target (killed vs. total casualties) in the final answer. \textbf{SIGHT}, guided by the \textbf{Information-Gain Hint}, maintains focus on the specific sub-metric requested, leading to a precise and correct final answer.}
\label{tab:case_study_3}
\begin{tabularx}{\linewidth}{lX}
\toprule
\multicolumn{2}{l}{\textbf{Question:} How many Union soldiers died at the battle of Gettysburg?} \\
\multicolumn{2}{l}{\textbf{Ground Truth:} \textcolor{correctgreen}{3,155}} \\
\midrule

\textbf{SIGHT} & \textcolor{taggray}{<think>} To determine the number of Union soldiers who died... we need to find information about the casualties... \textcolor{taggray}{</think>} \\
\textbf{(Correct)} & \textcolor{taggray}{<search>} how many union soldiers died at the battle of gettysburg \textcolor{taggray}{</search>} \\
 & \textcolor{taggray}{<result>} [Doc 1] ...Union casualties were 23,055 (\textcolor{evidenceblue}{\textbf{3,155 killed}}, 14,531 wounded, 5,369 captured or missing)... \textcolor{taggray}{</result>} \\
 & \textcolor{hintorange}{<hint> You have now found key information about one aspect of this question... you should consider the other aspect of this question. </hint>} \\
 & \textcolor{evidenceblue}{<self-evidence>} ...23,055 (3,155 killed, 14,531 wounded, 5,369 captured or missing)... \textcolor{evidenceblue}{</self-evidence>} \\
 & \textcolor{taggray}{<think>} From the search results, we can conclude that the number of Union soldiers who died... is 3,155. \textcolor{taggray}{</think>} \\
 & \textcolor{taggray}{<answer>} \textcolor{correctgreen}{\boxed{3,155}} \textcolor{taggray}{</answer>} \\

\midrule

\textbf{Baseline} & \textcolor{taggray}{<think>} To determine the number of Union soldiers who died... \textcolor{taggray}{</think>} \\
\textbf{(Incorrect)} & \textcolor{taggray}{<search>} how many union soldiers died at the battle of gettysburg \textcolor{taggray}{</search>} \\
 & \textcolor{taggray}{<result>} [Doc 1] ...were \textcolor{wrongred}{23,055} (3,155 killed, 14,531 wounded, 5,369 captured or missing)... \textcolor{taggray}{</result>} \\
 & \textcolor{taggray}{<self-evidence>} Battle of Gettysburg was fought July 1–3, 1863... The battle involved the largest number of casualties... \textcolor{taggray}{</self-evidence>} \\
 & \textcolor{taggray}{<think>} From the search results, we can see that the Union casualties... were 23,055, out of which 3,155 were killed. \textcolor{taggray}{</think>} \\
 & \textcolor{taggray}{<answer>} \textcolor{wrongred}{\boxed{23,055}} \textcolor{taggray}{</answer>} \\

\bottomrule
\end{tabularx}
\end{table*}

\begin{table*}[h]
\centering
\caption{Comprehensive case study demonstrating the synergistic effect of the three hint types in SIGHT. The \textbf{Information-Gain Hint} guides the task transition after the first hop; the \textbf{Reflection Hint} prompts gap analysis when the search results are perceived as insufficient; and the \textbf{De-duplication Hint} prevents redundant search loops, ensuring the model eventually extracts the precise evidence.}
\label{tab:triple_hint_case}
\begin{tabularx}{\linewidth}{lX}
\toprule
\multicolumn{2}{l}{\textbf{Question:} Determine the birth date of the director of the 2010 film \textit{Insidious}.} \\
\multicolumn{2}{l}{\textbf{Ground Truth:} \textcolor{correctgreen}{February 26, 1977}} \\
\midrule

\textbf{SIGHT} & \textcolor{taggray}{<think>} ...we need to first identify the director of the film. \textcolor{taggray}{</think>} \\
\textbf{(Triple Hint} & \textcolor{taggray}{<search>} 2010 American-Canadian supernatural horror film Insidious director \textcolor{taggray}{</search>} \\
\textbf{Synergy)} & \textcolor{taggray}{<result>} [Doc 1] \textit{Insidious} is a 2010 supernatural horror film directed by \textcolor{evidenceblue}{\textbf{James Wan}}... \textcolor{taggray}{</result>} \\
 & \textcolor{hintorange}{\textbf{<Information-Gain Hint>}} \textcolor{hintorange}{You have now found key information about one aspect... you should consider the other aspect of this question.} \\
 & \textcolor{evidenceblue}{<self-evidence>} ...directed by James Wan... released on April 1, 2011... \textcolor{evidenceblue}{</self-evidence>} \\
 & \textcolor{taggray}{<think>} Now that we know the director is James Wan, we can find out his birth date. \textcolor{taggray}{</think>} \\
 & \textcolor{taggray}{<search>} James Wan birth date \textcolor{taggray}{</search>} \\
 & \textcolor{taggray}{<result>} [Doc 2] James Wan (born \textcolor{evidenceblue}{\textbf{26 February 1977}}) is an Australian film director... \textcolor{taggray}{</result>} \\
 & \textcolor{hintorange}{\textbf{<Reflection Hint>}} \textcolor{hintorange}{This did not yield the answer. Analyze the gap... What is missing? Generate a new search query that targets the missing information.} \\
 & \textcolor{evidenceblue}{<self-evidence>} Wan was born in Kuching, Sarawak, Malaysia and is of Malaysian Chinese descent... \textcolor{evidenceblue}{</self-evidence>} \\
 & \textcolor{taggray}{<think>} ...still need to find his birth date. \textcolor{taggray}{</think>} \\
 & \textcolor{taggray}{<search>} James Wan birth date \textcolor{taggray}{</search>} \\
 & \textcolor{taggray}{<result>} \textit{[Empty Result]} \textcolor{taggray}{</result>} \\
 & \textcolor{hintorange}{\textbf{<De-duplication Hint>}} \textcolor{hintorange}{You have previously searched for 'James Wan birth date'. This did not yield the answer. You must generate a semantically distinct query now.} \\
 & \textcolor{evidenceblue}{<self-evidence>} ...Wan was born in Kuching... \textbf{James Wan was born on \textcolor{evidenceblue}{February 26, 1977.}} \textcolor{evidenceblue}{</self-evidence>} \\
 & \textcolor{taggray}{<answer>} \textcolor{correctgreen}{\boxed{\text{February 26, 1977}}} \textcolor{taggray}{</answer>} \\

\bottomrule
\end{tabularx}
\end{table*}

\end{document}